\documentclass[letterpaper]{article} 
\usepackage{aaai2026}  
\usepackage{times}  
\usepackage{helvet}  
\usepackage{courier}  
\usepackage[hyphens]{url}  
\usepackage{graphicx} 
\usepackage{natbib}  
\usepackage{caption} 
\usepackage{algorithm}
\usepackage{algorithmic}

\usepackage{newfloat}
\usepackage{listings}
\DeclareCaptionStyle{ruled}{labelfont=normalfont,labelsep=colon,strut=off} 
\floatstyle{ruled}
\newfloat{listing}{tb}{lst}{}
\floatname{listing}{Listing}
\usepackage{amssymb}
\usepackage{amsmath}
\usepackage{multirow}
\usepackage{booktabs}
\usepackage[table,xcdraw]{xcolor}
\usepackage{tabularx}
\usepackage{longtable}
\usepackage[most]{tcolorbox}

\newtcolorbox{blackpromptbox}[1][]{%
    breakable,
    colback=gray!10,
    colframe=black,
    fontupper=\small,
    left=0.5mm, right=0.5mm, top=1mm, bottom=1mm,
    boxrule=0.8pt,
    sharp corners,
    title={Prompt},
    fonttitle=\bfseries,
    #1 
}

\newtcolorbox{orangepromptbox}[1][]{%
    breakable,
    colback=orange!10!white,
    colframe=orange!70!black,
    fontupper=\small,
    left=0.5mm, right=0.5mm, top=1mm, bottom=1mm,
    boxrule=0.8pt,
    sharp corners,
    title={Prompt},
    fonttitle=\bfseries,
    #1 
}

\newtcolorbox{bluepromptbox}[1][]{%
    breakable,
    colback=blue!5!white,
    colframe=blue!50!black,
    fontupper=\small,
    left=0.5mm, right=0.5mm, top=1mm, bottom=1mm,
    boxrule=0.8pt,
    sharp corners,
    title={Prompt},
    fonttitle=\bfseries,
    #1 
}

\title{Mem-PAL: Towards Memory-based Personalized Dialogue Assistants for Long-term User-Agent Interaction}
\author {
    Zhaopei Huang\textsuperscript{\rm 1},
    Qifeng Dai\textsuperscript{\rm 2},
    Guozheng Wu\textsuperscript{\rm 1},
    Xiaopeng Wu\textsuperscript{\rm 2},
    Kehan Chen\textsuperscript{\rm 2},
    Chuan Yu\textsuperscript{\rm 2},\\
    Xubin Li\textsuperscript{\rm 2},
    Tiezheng Ge\textsuperscript{\rm 2},
    Wenxuan Wang\textsuperscript{\rm 1},
    Qin Jin\textsuperscript{\rm 1$\ast$}
}
\affiliations {
    \textsuperscript{\rm 1}Renmin University of China\\
    \textsuperscript{\rm 2}Taobao \& Tmall Group of Alibaba\\
    huangzhaopei@ruc.edu.cn, wuguozheng@ruc.edu.cn, qjin@ruc.edu.cn
}

\usepackage{bibentry}

\begin{document}

\maketitle

\renewcommand{\thefootnote}{\fnsymbol{footnote}} 
\footnotetext[1]{Qin Jin is the corresponding author.}
\renewcommand{\thefootnote}{\arabic{footnote}} 
\setcounter{footnote}{0}  

\begin{abstract}
With the rise of smart personal devices, service-oriented human-agent interactions have become increasingly prevalent. This trend highlights the need for personalized dialogue assistants that can understand user-specific traits to accurately interpret requirements and tailor responses to individual preferences. However, existing approaches often overlook the complexities of long-term interactions and fail to capture users' subjective characteristics. To address these gaps, we present \textbf{PAL-Bench}, a new benchmark designed to evaluate the personalization capabilities of service-oriented assistants in long-term user-agent interactions. In the absence of available real-world data, we develop a multi-step LLM-based synthesis pipeline, which is further verified and refined by human annotators. This process yields \textbf{PAL-Set}, the first Chinese dataset\footnote{We also provide an English-translated version.} 
comprising multi-session user logs and dialogue histories, which serves as the foundation for PAL-Bench. Furthermore, to improve personalized service-oriented interactions, we propose \textbf{H$^2$Memory}, a hierarchical and heterogeneous memory framework that incorporates retrieval-augmented generation to improve personalized response generation. Comprehensive experiments on both our PAL-Bench and an external dataset demonstrate the effectiveness of the proposed memory framework.
\end{abstract}

\begin{links}
    \link{Code, Dataset and Appendix}{https://github.com/hzp3517/Mem-PAL}
\end{links}

\section{Introduction}
\label{sec:intro}
\begin{figure}[!t]
\centerline{\includegraphics[width=1.0\columnwidth]{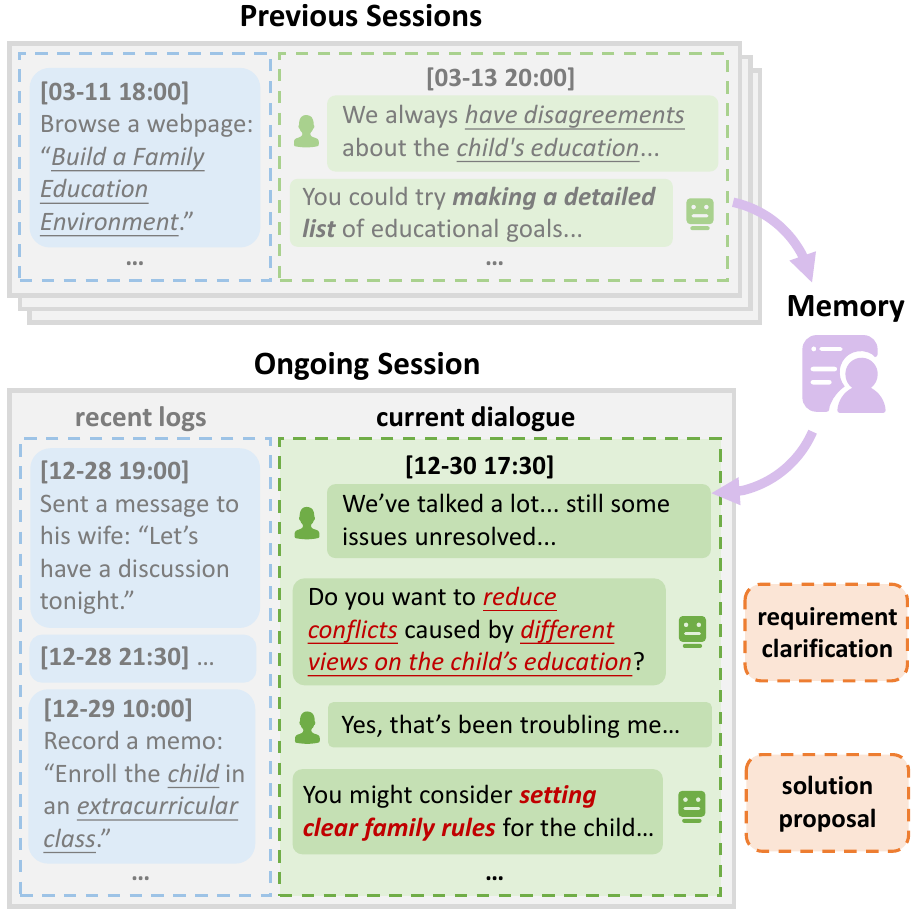}}
\caption{An example of our long-term, multi-session user-agent interaction data. The assistant is expected to leverage the historical interaction data (shown in lighter color) for memory modeling, enabling a more accurate understanding of user requirements and delivery of preference-aligned responses in the current dialogue.
}
\label{fig:intro}
\end{figure}

The development of mobile internet has significantly enhanced interactions between users and their personal smart devices, such as using smart bands to monitor health, sending messages via smartphones, or conversing with virtual assistants. We refer to this pattern of communication as \emph{user-agent interaction}, where the agent can access both the user's behavioral history and prior dialogue context. In such scenarios, an ideal service-oriented assistant should effectively leverage this interaction history for personalized modeling. This allows the assistant to deliver tailored solutions that align with individual user preferences and needs, without requiring cumbersome and repetitive explanations~\citep{ha2024clochat}, much like seeking help from a familiar pal.

Despite growing interest in service-oriented dialogue systems (e.g., medical assistants~\citep{zhang2023huatuogpt}, psychological counseling~\citep{zhang2024escot, zhang2024cpsycoun}, and mobile virtual assistants~\citep{guan2024intelligent}, etc.), most existing approaches treat users uniformly and lack the capacity to generate truly personalized responses. On the other hand, some recent studies have begun exploring long-term dialogue scenarios~\citep{xu2022msc, maharana2024locomo, wu2025longmemeval}, introducing benchmarks that focus on retrieving personal facts from long-term interactions. However, these efforts often overlook the more subjective and nuanced task of modeling user preferences and individualized requirements. Moreover, their histories are typically limited to dialogues, ignoring user behavioral records---an essential component of real-world user-agent interactions. To the best of our knowledge, only~\citet{wang2024emg-rag} have considered app screenshots as a form of behavioral history, but their dataset is not publicly available due to privacy concerns, which limits progress in developing and evaluating personalized service-oriented systems.

To address these challenges, we propose the first benchmark for personalized user-agent interaction, \textbf{PAL-Bench}. It introduces three evaluation tasks---\textit{Requirement Restatement}, \textit{Solution Proposal}, and \textit{Multi-turn Dialogue Interaction}---designed to evaluate the capability of service-oriented dialogue assistants to understand and adapt to users’ personalized requirements and preferences. 
Since collecting real-world long-term interaction data is costly and constrained by privacy concerns, we design a multi-stage, LLM-based data synthesis pipeline that incorporates verification and refinement procedures, resulting in a \textbf{PAL-Set}. This Chinese-language dataset contains both user behavior logs and user-assistant dialogues. 
As shown in Figure~\ref{fig:intro}, PAL-Set captures realistic long-term interaction patterns: it features 100 users, each with an average of 29 sessions, 996 behavioral logs, and 401 dialogue turns.
We also perform a human evaluation on the dataset, which confirms the high quality of the generated data.

To further support personalized modeling in these scenarios, we propose \textbf{H$^2$Memory}, a hierarchical and heterogeneous memory framework. In contrast to previous long-term memory modeling approaches~\citep{wang2025recursum, yuan2025conditionmem, zhong2024memorybank, ong2024theanine}, our approach explicitly models different forms of user history (e.g., behaviors versus dialogues) and introduces a two-level memory storage tailored to capture subjective user requirements and preferences, with update mechanisms for persona dynamics.
Equipped with H$^2$Memory and a retrieval-augmented generation (RAG) strategy, assistants are better positioned to serve personalized, context-aware responses, as required by PAL-Bench. We also validate the generalizability of our method on an external dataset.

Our contributions are threefold: 
(1) We present PAL-Bench, the first Chinese benchmark for long-term user-agent interactions supported by a scalable LLM-based data synthesis and human refinement pipeline. It features three tasks focused on modeling user personalized requirements and preferences.
(2) We propose H$^2$Memory, a hierarchical and heterogeneous memory framework that supports effective modeling, retrieval, and updating of diverse interaction histories for personalized service delivery.
(3) We demonstrate the quality of our PAL-Bench dataset through human evaluation, and validate the effectiveness of H$^2$Memory via comprehensive experimental analysis on both PAL-Bench and an external dataset, advancing research in personalized dialogue systems.

\section{PAL-Bench}
\label{sec:pal-bench}
\subsection{Dataset Construction}
\label{ssec:data_construction}
\begin{figure*}[!t]
\centerline{\includegraphics[width=1.0\textwidth]{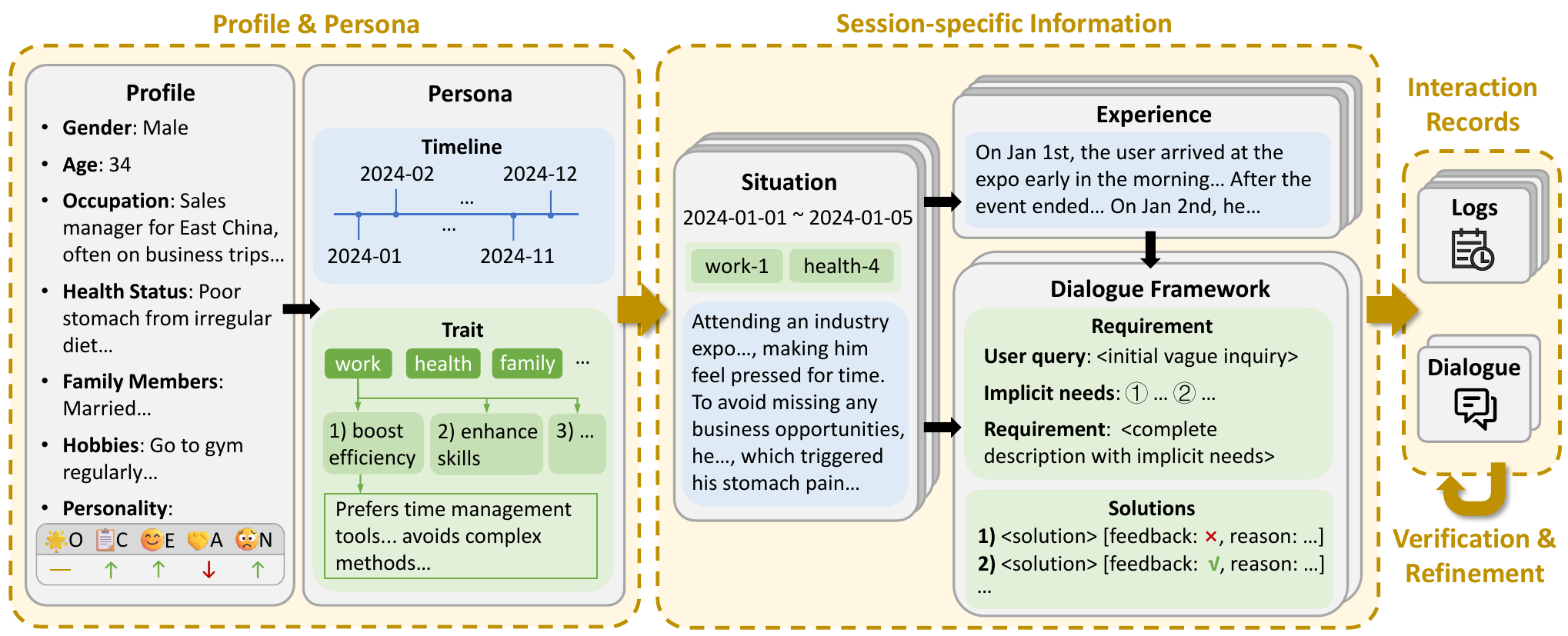}}
\caption{Overview of the generative pipeline for PAL-Set. We design a multi-stage LLM-based synthesis process to progressively specify the control information for interaction record generation. Additional verification and refinement steps are employed to ensure the final data quality.}
\label{fig:data_synthesis}
\end{figure*}

User interaction records over the long term often reflect stable user profiles and traits at the macro level, while exhibiting dynamic changes at finer granularity. To simulate such patterns, we design a multi-stage generation pipeline (Figure~\ref{fig:data_synthesis}). 
We first define each user's overall profile and create a corresponding persona. The persona is then expanded into multiple session-specific scenarios, which serve as fine-grained control signals for synthesizing interaction records. All synthesis steps are automatically performed using Qwen2.5-Max~\citep{qwen25}, followed by verification and refinement steps to ensure data quality.

\subsubsection{Profile and Persona}
We begin by generating a basic profile for each user, including gender, age, personality, and brief descriptions across four aspects: work, health, family, and leisure. 
Building on this profile, we further synthesize a persona that includes: (i) a personal timeline spanning several months, outlining monthly objective events to guarantee temporal coherence; and (ii) a set of user traits, covering multiple general requirement types with corresponding preference descriptions, to support subjective consistency over long-term interactions. 

\subsubsection{Session-specific Information}
{Since our ultimate goal is to synthesize multi-session interaction records, we need to expand the initial profile and persona into multiple session-specific pieces of information, which serve as references for the subsequent interaction records synthesis. We first expand the timeline for each month into multiple requirement-oriented situation entries. Each situation entry is a brief description composed of a few sentences, revolving around several requirement types predefined in the user's traits. We then expand each situation entry into a diary-style experience description that captures rich, detailed, and behaviorally grounded events over the same period. These experiences will help synthesize subsequent objective, time-stamped logs. For the guidance of synthesizing dialogues, we further construct a dialogue framework based on each situation.}

{Specifically, each dialogue framework may consist of multiple \emph{topics}, with each topic containing two parts: ``Requirements'' and ``Solutions''.}
The ``Requirements'' part includes three components: \emph{user query}, \emph{implicit needs}, and \emph{requirement}. The \emph{user query} represents the user's initial question for each topic, typically brief and underspecified to reflect real-world user behavior. The \emph{implicit needs} section includes two entries that are not explicitly stated in the query but are relevant to the user’s background or experiences and are expected to be inferred by the assistant. The \emph{requirement} section combines the user query and implicit needs described above to offer a complete description of the user's intent. For the ``Solutions'' part, we first generate 8 diverse candidate solutions that are all considered generally reasonable. Then we identify 2 positive solutions (highly aligned with the user's preferences) and 2 negative solutions (not aligned) among these candidates based on the predefined user profile and persona.

\subsubsection{Interaction Records}
{The long-term interaction records are the final output of our synthesis pipeline, comprising multi-session logs and dialogues that are accessible to the assistant.}
For log synthesis, we prompt the LLM to simulate {a variety of platforms on} a user's personal device and generate a series of records that indirectly reflect the {diverse} events experienced by the user.
For dialogue synthesis, we ensure that the generated dialogues adhere to the corresponding dialogue framework and reflect realistic flows commonly seen in service-oriented interactions. To achieve this, we define a set of dialogue actions for both user and assistant utterances. We also construct utterance-level dialogue templates that specify the action of each utterance and its linkage to specific components of the dialogue framework. {These templates can ensure that} the synthesized dialogues are consistent with our intended dialogue flows and content settings.

\subsubsection{Data Verification and Refinement}
To ensure the quality of data synthesis, we employ a series of data verification and refinement methods. For each LLM-based generation step, we define a specific output format and set up validation rules. The output is automatically checked after each generation, and a regeneration process would be triggered if the output does not meet the rules.
For the final logs and dialogues, we further perform validation and correction, targeting the following issues:
(1) Log entries that do not match the specified type;
(2) Dialogue utterances that are inconsistent with their assigned action;
(3) Content that contradicts the user’s profile or persona.

\subsection{Dataset Analysis}
\label{ssec:statistics}

\begin{table}[t]\small
\centering
\begin{tabular}{lll}
\toprule[1pt]
                              & History & Query \\ \midrule
Avg. \# sessions        & 25.7       & 3.3      \\
Avg. \# logs            & 888.7      & 107.5    \\
Avg. \# dialogue turns  & 361.7      & 39.3     \\
Avg. \# dialogue topics & 62.5       & 8.3      \\
Avg. \# months       & 8.4        & 1.0      \\ \bottomrule[1pt]
\end{tabular}
\caption{Statistics of PAL-Set. Each value represents the average across 100 users.}
\label{tab:dataset_statistics}
\end{table}

Our PAL-Set contains 100 synthetic users, each associated with long-term, multi-session interaction records. For each user, sessions are divided into a history set and a query set. The query set includes all sessions from the last month of interaction and serves as the evaluation samples, while the history set comprises all earlier sessions and is used solely as contextual input. Note that a query set may contain multiple sessions, and earlier sessions within the query set can also serve as part of the history when evaluating later sessions. 
Table~\ref{tab:dataset_statistics} summarizes the key statistics of PAL-Set.  Our dataset features a long interaction span (9.4 months on average) and a relatively large number of sessions (29 sessions) for each user, offering a rich long-term context. 

In addition, we conduct human evaluations to verify whether the generated logs and dialogues align with the predefined user profiles and personas. Annotators rated each sample on a scale from 1 (non-matching) to 3 (completely matching). The average scores were 2.75 for logs and 2.67 for dialogues, indicating that the synthetic data is highly consistent with the intended user characteristics, demonstrating the high quality of our dataset.

\subsection{Evaluation Tasks}
\label{ssec:evaluation_tasks}

Based on PAL-Set, we design three evaluation tasks as part of PAL-Bench: two single-turn question-answering tasks and one multi-turn dialogue interaction task. These tasks assess the assistant's ability to understand user requirements and generate personalized responses.

\noindent \textbf{1) Requirement Restatement.}
This task evaluates the assistant’s ability to accurately infer the user’s complete requirement from user histories in a single-turn QA setting. The input is the initial \emph{user query} for each topic in the query set, while the expected output is the corresponding complete \emph{requirement} description. We use BLEU score~\citep{papineni2002bleu} as the objective metric.
Since the reference \emph{requirement} descriptions are relatively abstract, we additionally introduce a GPT-4-based evaluation to specifically assess whether the generated content successfully captures the \emph{implicit needs} that are not explicitly stated in the \emph{initial query}, forming GPT-4 Scores for this task.

\noindent \textbf{2) Solution Proposal.}
This task evaluates the assistant’s ability to 
{understand user-specific preferences based on user histories and provide responses that meet these preferences}
in a single-turn QA format. 
It consists of two subtasks: ``solution generation'' and ``solution selection''. In ``solution generation'', the assistant is required to generate a solution description based on the given complete \emph{requirement}. 
{We evaluate the output with the BLEU score as an objective metric, taking the 2 predefined positive solutions in the dialogue framework as references.}
{In ``solution selection'', the assistant is additionally presented with the 8 candidate solutions and asked to identify the 2 positive ones. A Selection Score is then calculated based on the preference labels of the selected solutions.}

\noindent \textbf{3) Multi-turn Dialogue Interaction.}
Since our ultimate goal is to enhance the assistant’s performance in dialogue, we design evaluation tasks specifically targeting multi-turn interactions.  Considering the high cost of interacting with real users, we take advantage of the strong role-playing abilities of LLMs~\citep{chen2024role-play} to construct a User-LLM that simulates the predefined users in PAL-Set and interacts with the assistant. 
We also introduce an Evaluation-LLM responsible for automatically assessing the interaction quality of different memory modeling methods, which focuses on two key dimensions: requirement understanding (the assistant’s ability to accurately clarify the user's actual needs) and preference understanding (the extent to which the assistant’s proposed solutions align with user preferences).
We conduct pairwise comparative evaluations on both these two dimensions, and report the Win–Tie–Lose counts for our method over all evaluation samples. To reduce the effects of LLM randomness and positional bias on the evaluation, we follow the FairEval framework~\citep{wang2024faireval} to evaluate each pair 6 times with different input orders. In addition, we conduct human evaluations and analyze their correlation with LLM-based evaluations, as detailed in the Appendix.

\section{H$^2$Memory Framework}
\label{sec:memory_framework}
\begin{figure*}[!t]
\centerline{\includegraphics[width=0.9\textwidth]{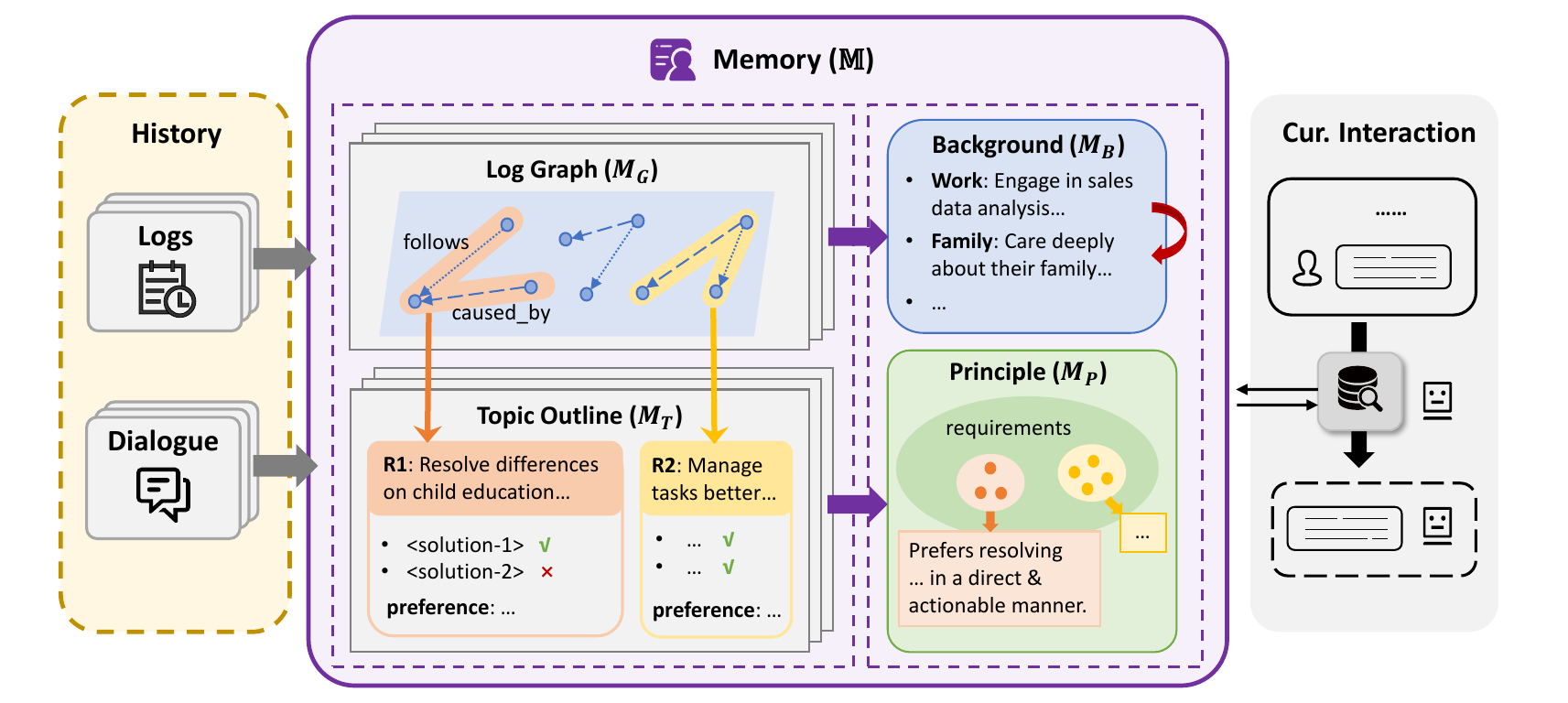}}
\caption{Overview of our method. We propose a hierarchical and heterogeneous memory mechanism (H$^2$Memory) to model user characteristics in user–agent interactions. Information from different sources is separately encoded into concrete- and abstract-level memory entries. The most relevant entries from each part are retrieved to enable personalized, retrieval-augmented response generation.
}
\label{fig:method}
\end{figure*}

\subsection{Problem Formulation}
\label{ssec:problem_formulation}
In our user-agent interaction scenario, each session $S$ consists of a series of logs $L = \{l_1, l_2, ..., l_m\}$ and a dialogue $D = \{u_1, a_1, ..., u_t, a_t\}$, where $l$ denotes a log record, $u$ and $a$ denote user and assistant utterances, respectively. 
Given the current session $S_c$, the logs collected between the end of $D_{c-1}$ and the beginning of the $D_c$ are taken as $L_c$. We define the interaction history $\mathbb H$ as all previous sessions along with the latest logs $L_c$: ${\mathbb H} = \{\{L_1, D_1\}, ..., \{L_{c-1}, D_{c-1}\}, L_c\}$. Let $Q$ denote the user's inquiry (along with preceding dialogue context, if any) in the current dialogue $D_c$, and $R$ denote the assistant's generated response. The task is formalized as: $R = {\mathrm {Assistant}}({\mathbb H}, Q)$.

However, due to the long-term nature of interactions, the original 
$\mathbb H$  can be prohibitively large and redundant, making it unsuitable as direct input. 
Therefore, we encode $\mathbb H$ into a concise and structured memory $\mathbb M$, from which relevant entries are retrieved to support personalized response generation.  As shown in Figure~\ref{fig:method}, our method integrates a hierarchical heterogeneous memory structure with retrieval-augmented generation (RAG).

\subsection{Memory Construction}
\label{ssec:memory_construction}
In user-agent interaction scenarios, both logs and dialogues exhibit substantial heterogeneity---in their information formats and in the user persona perspectives they convey. A single memory structure cannot effectively capture both aspects. 
To address this, we adopt a differentiated memory organization tailored to long-term histories.
For fragmented logs, we explicitly construct relational edges to form coherent and contextually complete situation descriptions. For service-oriented dialogues, which often follow dynamic topic flows, we define topic-level schemas and extract structured outlines to reflect the evolving conversation patterns.
Building on these specific memory entries, we further summarize and abstract the user’s overarching background and traits, resulting in a hierarchical two-tier memory structure.

\subsubsection{Log Graph}
\label{sssec:log_graphs}
Logs are fragmented observations reflecting user behavior. To construct a coherent understanding of user experiences over time, we explicitly model the relationships between log entries. Inspired by the commonsense knowledge graphs that connect knowledge nodes via various relation edges~\citep{hwang2021comet}, we instruct the LLM to identify relational edges (including the \textit{Caused\_by} and \textit{Follows} types) between logs in each session, forming relation set $\Upsilon = {\mathrm {LLM}}(L)$. These relations connect log entries chronologically, resulting in connected subgraphs $\{G_1, ..., G_g\}$.

We then instruct the LLM to generate a situation description $s_i$ for each subgraph $G_i$, integrating the logs through their relations. The situation entries corresponding to each subgraph constitute the first part of the memory in the $j$-th session, denoted $M_G^j = \{s_1^j, ..., s_g^j\}$.

\subsubsection{Background}
\label{sssec:background}
Since a single situation entry cannot comprehensively reflect all aspects of the user’s overall background, we construct a long-term background memory $M_B$ by summarizing across all situation entries in $\mathbb H$. We define several fixed aspects (e.g., work, family) and maintain a paragraph summary per aspect.
Given that the user's experiences continuously evolves over time, to ensure the background remains up-to-date, we adopt a recursive memory updating mechanism~\citep{wang2025recursum} for $M_B$. Given the $M_G^j$ of the $j$-th session, the updating process is: $M_B^{(j)} = {\mathrm {LLM}}(M_B^{(j-1)}, M_G^j)$. 

Additionally, we use the LLM to associate each situation entry $s_i$ in $M_G$ with one or more aspects in $M_B$, forming a multi-valued mapping ${\cal F}_{GB}: M_G \to {\cal P}(M_B)$, where ${\cal P}(M_B)$ is the power set of $M_B$.

\subsubsection{Topic Outline}
\label{sssec:topic_outlines}
A service-oriented dialogue may involve discussions on multiple topics, with each topic containing both the clarification of user requirements and the discussion of solutions. For each topic $t$ in the dialogue, we define an information schema $\{r, {\mathbf o}, p\}$. Here, $r$ represents the user requirements reflected in the topic, ${\mathbf o}$ includes the multiple solutions provided by the assistant along with the user's feedback on each solution, and $p$ summarizes the user's preference of solutions for the topic. The topic outline formed based on this schema can extract the key points of relevant topics from the original dialogue. Therefore, we guide the LLM to segment each dialogue $D$ into $\tau$ topic-specific parts and generate a set of such outlines $T$, where $T = \{t_1, ..., t_{\tau}\}$.
However, users often express their requirements in dialogues quite briefly, lacking the background or situation, which are often reflected in the logs from the corresponding period. To effectively integrate information from both sources, we retrieve $k$ situation entries ${\mathbf s}$ from $M_G^j$ of the corresponding $j$-th session that are most relevant to the requirement $r$. Based on this, we rewrite the $r$ into a more detailed description $\hat{r}$, i.e., $\hat{r} = {\mathrm {LLM}}(r, {\cal R}({\mathbf s}))$, where ${\cal R}({\mathbf s})$ denotes the retrieved situation entries related to $r$. We perform requirement rewriting for each $r$ in $T$ to get $\hat{T}$, and add these outlines to the memory bank, i.e., $M_T^{(j)} = M_T^{(j-1)} \oplus \hat{T}$.

\subsubsection{Principle}
\label{sssec:principle}
We further abstract several overall requirement types and corresponding preference principles from numerous specific topic outlines to form the memory $M_P$. To be specific, we first initialize $M_P$ using all dialogues in the history set. All requirements (denoted as ${\mathbf r}$) extracted from these dialogues are encoded as features, and then we apply the KMeans clustering algorithm to obtain $n$ clusters. For each cluster $C_i$, we instruct the LLM to extract the requirement type $\gamma_i$ and preference principle $\rho_i$ from bunch of specific descriptions ($\mathbf{r}_i$ \& $\mathbf{p}_i$) belonging to the $C_i$. Thus, we obtain $M_P = \{\mathbf{\gamma}, \mathbf{\rho}\}$.
Note that we retain the cluster assignments of all specific requirements, thereby forming a mapping ${\cal F}_{TP}: M_T \to M_P$.

Considering that the query set may contain multiple sessions, $M_P$ needs to be continuously updated for new query sessions. For a requirement entry $r$, we first find the closest cluster center $C_i$, then update $C_i$ with the features of $r$, and send both the previous $\gamma_i$ and the current $r$ to the LLM, prompting it to update the $\gamma_i$ if necessary. We perform a similar update for the corresponding $\rho_i$ as well.

\begin{table*}[!t]\small
\centering
\begin{tabular}{l|ccccc|ccccc}
\toprule[1pt]
\multirow{2}{*}{Methods} & \multicolumn{5}{c|}{Requirement Restatement}                                     & \multicolumn{5}{c}{Solution Proposal}                                            \\
                         & B-1         & B-2         & B-3        & B-4        & G-Score    & B-1         & B-2        & B-3        & B-4        & S-Score \\ \midrule
Vanilla (w/o log)        & 13.59          & 5.76           & 2.58          & 1.41          & 17.50          & 18.85          & 6.96          & 3.52          & 2.06          & 18.95           \\
Vanilla (with log)       & 19.71          & 8.85           & 4.10          & 2.29          & 23.00          & 19.76          & 7.68          & 4.00          & 2.32          & 22.88           \\ \midrule
Turn-level RAG           & 22.74          & 10.54          & 4.94          & 2.69          & 26.85          & 19.15          & 7.43          & 3.89          & 2.29          & 24.09           \\
Session-level RAG        & 23.81          & 11.24          & 5.42          & 3.06          & 29.33          & 19.66          & 7.80          & 4.00          & 2.33          & 33.78           \\ \midrule
RecurSum~\citep{wang2025recursum}                 & 23.29          & 10.64          & 4.95          & 2.75          & 28.36          & 19.89          & 7.59          & 3.96          & 2.26          & 25.61           \\
ConditionMem~\citep{yuan2025conditionmem}             & 23.31          & 10.42          & 4.86          & 2.66          & 27.78          & 19.42          & 7.42          & 3.85          & 2.22          & 25.49           \\
MemoryBank~\citep{zhong2024memorybank}               & 23.89          & 11.11          & 5.23          & 2.91          & 28.57          & 20.49          & 8.12          & 4.07          & 2.34          & 29.85           \\ \midrule
\textbf{H$^2$Memory (ours)}     & \textbf{26.67} & \textbf{12.18} & \textbf{5.68} & \textbf{3.09} & \textbf{32.54} & \textbf{22.24} & \textbf{8.38} & \textbf{4.39} & \textbf{2.65} & \textbf{38.32}  \\
H$^2$Memory (w/o $M_G$)                  & 26.32          & 11.93          & 5.42          & 2.93          & 30.90          & 22.19          & 8.24          & 4.35          & 2.62          & 37.71           \\
H$^2$Memory (w/o $M_B$)                  & 26.22          & 11.95          & 5.53          & 3.05          & 31.30          & 21.84          & 8.17          & 4.19          & 2.46          & 36.69           \\
H$^2$Memory (w/o $M_T$)                  & 24.04          & 11.01          & 5.20          & 2.82          & 28.00          & 18.23          & 6.49          & 3.21          & 1.86          & 28.09           \\
H$^2$Memory (w/o $M_P$)                  & 26.33          & 11.90          & 5.44          & 2.90          & 31.51          & 21.97          & 8.16          & 4.23          & 2.49          & 36.26           \\ \bottomrule[1pt]
\end{tabular}
\caption{Performance of the requirement restatement and solution proposal tasks. B-1 to B-4 represent BLEU scores. G-Score and S-Score denote the GPT-4 Score and the Selection Score described in ``Evaluation Tasks'', respectively.}
\label{tab:single_turn_result}
\end{table*}

\subsection{Memory-based RAG}
\label{ssec:memory-based_rag}
The final hierarchical and heterogeneous memory structure is ${\mathbb M} = \{M_G, M_B, M_T, M_P\}$, where $M_G$ and $M_T$ encode concrete-level information extracted from logs and dialogue context, respectively, while $M_B$ and $M_P$ capture abstract-level information. For an inquiry $Q$ in the current $j$-th session, we retrieve relevant entries from each part of ${\mathbb M}$ as personalized context.
Specifically, we first retrieve the $k$ most relevant entries from the recent situation memory $M_G^j$, and then find the corresponding long-term background from $M_B$ for the aspects associated with these situations. Next, we retrieve the $k$ most relevant entries from the memory bank of the entire $M_T$ part, since we consider that user traits are more stable factors in the long term compared to situations, and we further find the abstract principles in $M_P$ corresponding to the specific topic entries. The whole retrieval process can be represented as: 
\begin{equation*}
\begin{aligned}
  \mathbf{m}\, =\;&  \{{\cal R}(M_G^j), {\cal F}_{GB}({\cal R}(M_G^j)), {\cal R}(M_T^{(j)}), \\
  & {\cal F}_{TP}({\cal R}(M_T^{(j)}))\}
\end{aligned}
\end{equation*}
where $\mathbf{m}$ denotes the retrieved personalized information, enabling us to realize a personalized augmented response generation process $R=\mathrm{LLM}(\mathbf{m}, Q)$.

\section{Experiments}
\label{sec:experiments}
\subsection{Experimental Setting}
\label{ssec:experimental_setting}

\subsubsection{Implementation Details}
\label{sssec:implementation_details}
For retrieval, we use the ``paraphrase-multilingual-mpnet-base-v2''~\citep{reimers2019sbert, song2020mpnet} as the encoder to extract text features from memory entries and queries, retrieving the top $k=3$ most similar memory entries each time by cosine similarity.
Qwen-Max-0428 is the base model (i.e., the Assistant-LLM) for memory construction and response generation in all experiments.
In the multi-turn dialogue interaction task, User-LLM and Evaluation-LLM are Qwen2.5-Max and GPT-4-turbo, respectively.

\subsubsection{Compared Baselines}
\label{sssec:baseline_and_comparison_methods}
We compare our method with several baseline approaches using the same base model.
\textit{Vanilla (w/o log)} refers to using only the current query as input without any interaction history, while \textit{Vanilla (with log)} includes the logs from the current session but still excludes previous sessions. \textit{Turn-level RAG} and \textit{Session-level RAG} retrieve historical dialogues at different granularities (utterance-level and session-level), respectively, while also incorporating logs from the current session. 
We also reimplement three prior memory-based methods: \emph{RecurSum}\citep{wang2025recursum}, \emph{ConditionMem}\citep{yuan2025conditionmem}, and \emph{MemoryBank}\citep{zhong2024memorybank}. As these methods did not originally account for logs in PAL-Bench, we add basic log processing for a fairer comparison.
Furthermore, all retrieval-based baselines follow the same retrieval settings as our method.

\subsection{Experimental Results on PAL-Bench}
\label{ssec:experimental_results}

\subsubsection{1) Requirement Restatement}
\label{sssec:requirement_restatement}
The left part of Table~\ref{tab:single_turn_result} presents the experimental results for the requirement restatement task.
Incorporating logs under the vanilla setting improves performance, confirming that the logs in our PAL-Set are meaningfully related to user requirements. Further gains are observed when incorporating long-term historical information, suggesting that PAL-Set effectively captures the consistency of users’ intrinsic characteristics over time. Our proposed H$^2$Memory framework achieves the best performance among all baselines, demonstrating its effectiveness in understanding requirements.

We also perform an ablation study on the four components of our memory structure.  
Among them, the relevant requirement descriptions in $M_T$ extracted from historical dialogues provide the most significant contribution to understanding the current user needs.
Nonetheless, the other memory components also yield measurable benefits, highlighting the complementary nature of the full memory design.

\subsubsection{2) Solution Proposal}
\label{sssec:solution_proposal}
The right part of Table~\ref{tab:single_turn_result} presents the experimental results for the solution proposal task. The overall trends mirror those observed in the requirement restatement task: the constructed long-term interaction records in our PAL-Set can effectively reflect specific user preferences, and our proposed method excels at modeling such preferences. Notably, our approach achieves a significant advantage over all baselines, particularly in the selection score. This indicates that user preferences tend to be abstract and nuanced, requiring more sophisticated modeling strategies beyond simple fact extraction from interaction history. Additionally, the ablation results reveal: removing $M_B$ (background memory) leads to a larger performance drop than removing $M_G$ (situation memory), which is the reverse of the trend observed in the requirement restatement task. This highlights that user preferences are more stable, high-level traits that accumulate over time, whereas requirements are more contextually grounded in recent events.

\subsubsection{3) Multi-turn Dialogue Interaction}
\label{sssec:dialogue_interaction}

\begin{table}[!t]\small
\centering
\begin{tabular}{lcc}
\toprule[1pt]
Ours vs Baseline   & Requirement & Preference \\ \midrule
Vanilla (w/o log)  & \textbf{478} / 29 / 319  & \textbf{480} / 18 / 328 \\
Vanilla (with log) & \textbf{447} / 33 / 346  & \textbf{452} / 22 / 352 \\
RecurSum           & \textbf{421} / 29 / 376  & \textbf{439} / 18 / 369 \\
ConditionMem       & \textbf{396} / 42 / 388  & \textbf{413} / 19 / 394 \\
MemoryBank         & \textbf{449} / 33 / 344  & \textbf{452} / 25 / 349 \\ \bottomrule[1pt]
\end{tabular}
\caption{Performance of the multi-turn dialogue interaction task. We report the ``Win/Tie/Lose'' numbers of our method compared to other baseline methods across all query topics.}
\label{tab:multi_turn_result}
\end{table}

Table~\ref{tab:multi_turn_result} presents the “Win/Tie/Lose” statistics of our method in comparison with multiple baseline methods on the multi-turn dialogue interaction task. The results indicate that our method exhibits the most significant advantage over the vanilla baseline, which does not incorporate any interaction history. Moreover, our method consistently outperforms other memory-based methods across both requirement and preference understanding dimensions.
These findings suggest that our proposed dual-level heterogeneous memory modeling mechanism is effective in multi-turn dialogue settings and holds promise for improving user satisfaction in interactions.
A qualitative case study is provided in the Appendix to further illustrate these outcomes.

While our method achieves strong overall performance, it still incurs a non-negligible number of “Lose” cases in the comparison with the vanilla baseline. 
We consider the primary reason is that, although the User-LLM follows predefined personas and dialogue actions, its utterances retain some randomness, which may also be a factor that guides Assistant-LLM responses toward varying content and affects their evaluation.
Nonetheless, despite the existence of this factor, the overall trend clearly supports the effectiveness of our method in improving multi-turn dialogue interaction.

\subsection{Validation on External Dataset}
\label{ssec:validation_on_external_dataset}

\begin{table}[!t]\small
\centering
\begin{tabular}{lc}
\toprule[1pt]
Methods            & Accuracy \\ \midrule
Vanilla            & 10.00    \\
RecurSum~\citep{wang2025recursum}           & 10.00    \\
ConditionMem~\citep{yuan2025conditionmem}       & 40.00    \\
MemoryBank~\citep{zhong2024memorybank}         & 23.33    \\ \midrule
\textbf{Ours ($M_T + M_P$)} & \textbf{50.00}    \\
Ours ($M_T$)       & 40.00    \\
Ours ($M_P$)       & 46.67    \\
\bottomrule[1pt]
\end{tabular}
\caption{Experimental results on the ``single-session-preference'' subset in LongMemEval~\citep{wu2025longmemeval}.}
\label{tab:longmemeval_validation}
\end{table}

{To verify the generalizability of our method, we also conduct experiments on the ``single-session-preference'' subset of LongMemEval~\citep{wu2025longmemeval} since its scenario is most similar to our PAL-Bench, with a focus on understanding user preferences. 
Besides, this dataset is in English, which can evaluate the applicability of our method to different languages.
However, their data only contains dialogues without logs. Therefore, we only employ the dialogue modeling component ($M_T$ + $M_P$) of our proposed H$^2$Memory framework in this experiment.}

As shown in Table~\ref{tab:longmemeval_validation}, combining both specific-level memory ($M_T$) and abstract-level memory ($M_P$) achieves the best performance and outperforms other approaches, demonstrating the generalizability of our method on external data beyond our PAL-Bench.

\section{Related Work}
\label{sec:realted_work}
\noindent \textbf{Long-term Dialogue Benchmarks.}
Recently, several long-term dialogue benchmarks have been proposed to facilitate research on personalized dialogue systems. \citet{xu2022msc} construct a multi-session dialogue dataset MSC by extending the Persona-Chat~\citep{zhang2018persona-chat}. ~\citet{jang2023cc} build a multi-session dialogue dataset CC, which includes relationships between dialogue roles. In addition, SHARE~\citep{kim2024share} and LoCoMo~\citep{maharana2024locomo} construct long-term dialogue datasets featuring shared memory and multi-modal histories, respectively. While the above works mainly focus on human-human interactions, some benchmarks also address user-assistant interactions and evaluate different aspects of assistant capabilities. 
{LongMemEval~\citep{wu2025longmemeval} mainly focuses on the extraction and recall of user facts from dialogue history, with only a small subset addressing user preferences.}
ImplexConv~\citep{li2025implexconv} centers on implicit reasoning over subtle information, and MapDia~\citep{wu2025proactive_dialogue} aims to enable proactive topic-shifting by assistants. 
{However, these works overlook the subjective characteristics of users,}  
and they do not consider user behavior histories in user-agent interaction settings. These are the main differences from our proposed PAL-Bench.

\noindent \textbf{Personalized Response Generation Methods.}
Personalized response generation methods fall into three categories. The first directly includes all user history in the prompt, but this only works for models with long-context support and often misses key personalized details.
The second type is memory parameterization~\citep{ma2021dhap,zhang2024milp,zhang2024malp,liu2024pplug}, but such methods cannot explicitly organize memory entries to handle complex personalized scenarios, and they are unsuitable for API-based LLMs since parameter fine-tuning is needed. The third type constructs external memories and employs retrieval-augmented generation (RAG) methods to enhance generation~\citep{lu2023memochat, zhong2024memorybank, wang2024emg-rag, yuan2025conditionmem}. Our work also falls into this type, but unlike previous studies, we design a hierarchical and heterogeneous memory structure for service-oriented user-agent interaction scenarios.

\section{Conclusion}
In this work, we focus on personalized, long-term, service-oriented interactions. To support research in this area, we design a multi-stage data synthesis pipeline and construct the first Chinese user-agent long-term interaction dataset, PAL-Set. Building on this, we introduce a new evaluation benchmark, PAL-Bench, aimed at assessing assistants’ abilities to understand user requirements and preferences based on long-term interaction histories.
Additionally, we propose a hierarchical and heterogeneous memory modeling framework, H$^2$Memory, to improve response generation in personalized dialogue settings.
Experimental results demonstrate the effectiveness of our proposed memory framework.

\appendix
\section*{Acknowledgments}
This work was sponsored by CCF-ALIMAMA TECH Kangaroo Fund (NO. CCF-ALIMAMA OF 2024007).

\bibliography{refs}

\section{Ethical Statements}
In this paper, we employ LLMs to simulate user-agent interactions. We take care to ensure that the generated data is free from harmful, biased, offensive, or pornographic content. Moreover, all user profiles are entirely fictional, eliminating concerns related to the disclosure of real users’ identities, addresses, or contact information.

\section{Details of PAL-Bench}
\label{sec:PAL-Bench_details}
\subsection{Pipeline Details}
\label{ssec:pipeline_details}
Here, we provide more details for each step of the data synthesis process.

\subsubsection{Profile}
The brief descriptions of users in the domains of work, health, family, and leisure are each composed of 2 to 3 sentences. User personality is described according to the Big Five personality model~\citep{mccrae1992big-five-personality}, which consists of five dimensions: openness, conscientiousness, extroversion, agreeableness, and neuroticism. For each dimension, we instruct the LLM to assign a value of ``high'', ``medium'', or ``low''. The inclusion of personality information is motivated by the consideration that user preferences can be strongly associated with their personality traits.

\subsubsection{Persona}
For each month in the user timeline, a minimum of five sentences is required to describe various objective events the user may encounter. The user’s requirement types should be summarized as simple, abstract phrases that reflect overall patterns in the user’s long-term experiences, rather than incidental occurrences. For each requirement type, both positive and negative preferences should be described based on the user’s identity, experiences, and personality. Each preference description should be 4 to 5 sentences.

\subsubsection{Situation}
We expand the timeline description for each month into 4 to 6 situations. Each situation is described in at least 5 sentences and revolves around 1 to 3 requirement types. When generating the specific situations for each month, all previously expanded situations from the prior month (i.e., the month just before the current one) are additionally provided as input to avoid temporal inconsistencies in situation details.

\subsubsection{Experience}
To ensure temporal consistency, the prior situation is provided as input when expanding a situation entry into a detailed experience description.

\subsubsection{Dialogue Framework}
For the ``Solutions'' part, we adopt a two-stage generation approach. First, without providing any user-specific information, we ask the LLM to generate 8 diverse candidate solutions for each \emph{requirement}. Then, after providing the user's profile and preferences as input, we ask the LLM to select the 2 solutions that best match and the 2 that least match the user's personalized preferences from the candidate solutions. The 4 selected solutions can then be used for dialogue synthesis, while the full set of 8 candidate solutions can be used for the solution proposal task. This approach ensures that negative solutions are still generally relevant to the current requirement. 
Based on our prior exploration, if we directly ask the LLM to generate negative solutions that do not match the user's preferences in a single step, these solutions often conflict with the user requirements, which diminishes the significance of personalized preference modeling.

\subsubsection{Logs}

{
\renewcommand{\arraystretch}{1.3}
\begin{table*}[!t]\small
\centering
\begin{tabularx}{\textwidth}{lX}
\toprule[1pt]
Log Type & Referenced Content Format \\ \midrule
Web Search         & ``The user searched for \textless keywords:string\textgreater  and viewed \textless content summary\textgreater .''              \\
Content Publishing         & ``The user published a \textless type: post/article/video\textgreater  on \textless online platform:string\textgreater , with the main content summarized as: \textless one-sentence summary\textgreater .''               \\
Content Browsing         & ``The user browsed a \textless type: article/product page/video\textgreater  on \textless online platform:string\textgreater , with the main content summarized as: \textless one-sentence summary\textgreater .''               \\
Message Sending         & ``The user sent a message to contact:string: `\textless message content\textgreater '.''               \\
Message Receiving         & ``The user received a message from \textless contact/device platform:string\textgreater : `\textless message content\textgreater '.''               \\
Schedule Management         & ``The user created a schedule named `\textless schedule name\textgreater ' in system:string, time: \textless start time\textgreater  - \textless end time\textgreater , location: \textless specific location\textgreater .''               \\
Transaction Record         & ``The user completed a purchase of \textless type: product/service\textgreater  via channel:string, product: \textless name + specification\textgreater .''               \\
Device Operation         & ``The user performed a \textless operation type\textgreater  operation on \textless device type\textgreater : \textless specific operation details (may include specific data or numbers)\textgreater .''               \\ \bottomrule[1pt]
\end{tabularx}
\caption{Predefined log types and referenced formats.}
\label{tab:log_type_definition}
\end{table*}
}

{
\renewcommand{\arraystretch}{1.3}
\begin{table*}[!t]\small
\centering
\begin{tabularx}{\textwidth}{llX}
\toprule[1pt]
Timestamp & Log Type & Log Content \\ \midrule
2024-08-13 07:30 & Device Operation & The user set up navigation on their smartphone, with the destination set to 15th floor, Guanghui Building. \\
2024-08-13 08:00 & Device Operation & The user logged into the market research system on their office computer and viewed the latest data reports. \\
2024-08-19 19:00 & Device Operation & The user browsed a live course on an online education platform, titled ``Practical Paths to Transition from Traditional Roles to Emerging Fields''.
 \\ \bottomrule[1pt]
\end{tabularx}
\caption{Several logs of the user ``0095'' in our dataset. Although these logs all belong to the ``Device Operation'' type, they reflect various types of actions performed by the user on different platforms.}
\label{tab:log_case}
\end{table*}
}

For log synthesis, we predefine 8 types of logs and their corresponding referenced formats, as shown in Table~\ref{tab:log_type_definition}. Each generated log entry should belong to one of these types. 
Note that our log formats are abstract and flexible, enabling them to accommodate a wide range of user behaviors. For example, as shown in Table~\ref{tab:log_case}, the synthesized user performs diverse actions on multiple platforms even under the same ``Device Operation'' log type. These include navigating on a mobile phone, using the office system on a computer and tracking fitness with a smart band. Therefore, the predefined log formats can preserve the diversity of user behaviors well.

\subsubsection{Dialogues}

{
\renewcommand{\arraystretch}{1.3}
\begin{table*}[!t]\small
\centering
\begin{tabularx}{\textwidth}{lX}
\toprule[1pt]
Dialogue Action              & Description            \\ \midrule
\multicolumn{2}{c}{\cellcolor[HTML]{EFEFEF}User}      \\
Topic Inquiry & The user initiates an inquiry to the assistant about a certain topic, which is usually brief and somewhat vague. \\
Requirement Confirmation & The user effectively responds to the assistant’s requirement inference, confirming their actual requirement. \\
Solution Discussion & In response to a proposal provided by the assistant, the user does not give a clear positive or negative evaluation of the overall proposal, but instead discusses specific details of the proposal with the assistant. \\
Solution Feedback  & The user expresses a clear positive approval or negative disapproval attitude toward the proposal suggested by the assistant. \\
\multicolumn{2}{c}{\cellcolor[HTML]{EFEFEF}Assistant} \\
Requirement Prediction & The assistant proactively infers the user's implicit requirement behind their inquiry, based on the user's profile or relevant experience. \\
Solution Proposal & The assistant proposes a solution as a suggestion for the user, based on the user's specific requirement under the current topic. \\
Solution Discussion  & The assistant responds to the user's opinions or inquiries about the current proposal, further discussing the proposal with the user based on the user’s profile or relevant experience, and tries to persuade the user to accept the proposal. \\
Feedback Response & The assistant responds to the user's expressed positive or negative attitude. It is important to note that no other solutions outside those specified in the dialogue framework should be offered here. In particular, when the user expresses a negative attitude, the assistant should only express apology or regret, and should not attempt to recommend other solutions to the user. \\ \bottomrule[1pt]
\end{tabularx}
\caption{Predefined dialogue actions of the user and the assistant.}
\label{tab:dialogue_action_definition}
\end{table*}
}

\begin{table*}[!t]
\centerline{\includegraphics[width=0.95\textwidth]{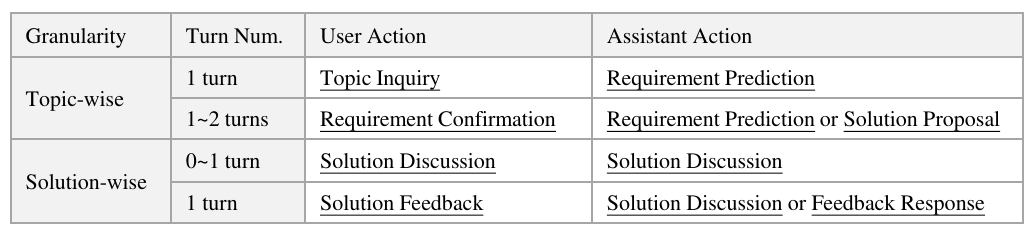}}
\caption{Predefined user and assistant dialogue patterns for our dialogue synthesis, where the assistant's dialogue action in the same turn depends on the user's dialogue action.}
\label{tab:dialogue_action_dynamic}
\end{table*}

For dialogue synthesis, we predefine 4 dialogue actions each for the user and the assistant, tailored to the service-oriented dialogue scenario (see Table~\ref{tab:dialogue_action_definition}). Additionally, we define the transition patterns between these actions to simulate realistic user-assistant interaction dynamics, as shown in Table~\ref{tab:dialogue_action_dynamic}. Specifically, the assistant first clarifies the user's initial query. Once the user confirms their needs, the assistant provides concrete solutions. The dialogue may then continue with the user and assistant discussing these solutions further or with the assistant responding to the user's feedback on the proposed solutions.

\begin{table*}[!t]\small
\centering
\begin{tabular}{lcccc}
\toprule[1pt]
\textbf{Dataset}        & \textbf{\begin{tabular}[c]{@{}c@{}}Avg. Sessions/Turns\\ per User\end{tabular}} & \textbf{Time Span} & \textbf{\begin{tabular}[c]{@{}c@{}}Construction\\ Method\end{tabular}} & \textbf{\begin{tabular}[c]{@{}c@{}}Interaction\\ Scenario\end{tabular}} \\ \midrule
PersonaChat~\citep{zhang2018persona-chat}             & 1 / 14.8                                                               & -                      & Crowdsourcing                                                 & human-human                                                    \\
MSC~\citep{xu2022msc}                     & 5 / 53.3                                                               & a few days             & Crowdsourcing                                                 & human-human                                                    \\
CC~\citep{jang2023cc}                      & 5 / 58.5                                                               & a few hours - years    & LLM-generated                                                 & human-human                                                    \\
LoCoMo~\citep{maharana2024locomo}                  & 19.3 / 304.9                                                           & a few months           & LLM-generated                                                 & human-human                                                    \\
LongMemEval~\cite{wu2025longmemeval}             & 500 / 5K                                                               & a few months           & LLM-generated*                                                & user-assistant                                                 \\
\textbf{PAL-Set (ours)} & 28.9 / 400.9                                                           & a few months           & LLM-generated                                                 & user-agent                                                     \\ \bottomrule[1pt]
\end{tabular}
\caption{Comparison of PAL-Set with existing datasets. * denotes that a large amount of unrelated dialogue data from publicly released user-AI datasets was added in addition to LLM-generated dialogues, resulting in much longer interaction histories.}
\label{tab:dataset_comparison}
\end{table*}

\subsection{Comparison with Other Datasets}
\label{ssec:datasets_comparison}
Table~\ref{tab:dataset_comparison} presents a comparison between our dataset and other existing datasets. The previous datasets are all limited to long-term dialogue scenarios, while our PAL-Set focuses on long-term user–agent interactions that include both dialogue and log records. Besides, each user in our dataset has a longer interaction history (an average of 28.9 sessions and 400.9 dialogue turns) than most other datasets, with the only exception being LongMemEval~\citep{wu2025longmemeval}. However, it should be emphasized that in LongMemEval, each user actually has very few persona-grounded dialogues, with most additional dialogues sourced from ShareGPT~\citep{zheng2023sharegpt} and UltraChat~\citep{ding2023ultrachat} that are unrelated to the current user, simply to extend the context length. Although such a setting can be used to evaluate the ``needle-in-a-haystack'' capability of LLMs, we claim that it is inadequate to evaluate the ability to integrate and abstract long-term behaviors for the same user. Therefore, in our PAL-Set, we do not simply extend the interaction history in this way. Instead, we simulate a person’s intrinsically consistent long-term behaviors and interaction dynamics through a carefully designed multi-step data construction process.

\subsection{Human Evaluation of Data Quality}
\label{ssec:dataset_human_evaluation}
We conduct a human evaluation of the constructed PAL-Set to assess the quality of the synthesized logs and dialogues. Specifically, we randomly sample 5 users from the dataset and select 10 sessions from each user's interaction records, resulting in a total of 50 sessions. We recruit three students majoring in computer science and provide them with reasonable compensation. They are asked to evaluate the logs and dialogues in these selected sessions. For the logs, annotators are provided with the user's profile and the corresponding user experiences, and are asked to judge whether the generated logs reflected the key events in the given situation. For the dialogues, annotators are given the user's profile as well as the predefined overall requirements and preferences, and are required to determine whether the user's utterances in the dialogues contradicted the predefined character traits.
For each session, annotators are asked to assign a score ranging from 1 to 3, where 1 indicates the generated content did not match the predefined content, 2 indicates a basic match, and 3 indicates a complete match.

The evaluation results show that both logs and dialogues rarely receive a score of 1. The average score given by the three annotators was 2.75 for logs and 2.67 for dialogues. This demonstrates that our proposed data synthesis process can successfully generate personalized and controllable interaction records, resulting in a high-quality personalized long-term interaction dataset.

\subsection{Dataset License}
The PAL-Set will be released under the CC BY-NC 4.0 license\footnote{https://creativecommons.org/licenses/by-nc/4.0/} and is intended for research purposes only.

\subsection{Details of Evaluation Setup}
\label{ssec:evaluation_setup_details}

\subsubsection{Requirement Restatement}
We use the ``Requirements'' part of the ``Dialogue Framework'' generated during data construction as both input and reference output for the task. The input is the initial vague inquiry (\emph{user query}) corresponding to each topic, while the ground-truth for BLEU score\footnote{We first use the jieba library to segment the original Chinese text, and then calculate the BLEU scores using the sentence\_bleu function from the NLTK library.} calculation is the user’s complete \emph{requirement} description from the framework. The GPT-4 Score focuses on the degree to which the output matches the \emph{implicit needs} specified in the framework. Since the predefined \emph{implicit needs} include two items, we ask GPT-4 (GPT-4o-0513) to assign a score from 0 to 2 for each sample, adding 1 point for each matched item and 0.5 points if only partially matched. To align with the range of BLEU score, we scale the final scores to the [0, 100] interval for reporting.

\subsubsection{Solution Proposal}
We utilize the ``Solutions'' part of the ``Dialogue Framework'' in this task. For the ``solution generation'' subtask, we use the positive solutions from the framework as references to calculate the BLEU score. Since each topic corresponds to two positive solutions, for each sample we take the maximum score computed with the two references.
For the ``solution selection'' subtask, we acquire the scores directly based on predefined polarity labels of the selected solutions. Among the candidate solutions, two are considered positive, two are considered negative, and the remaining four are considered neutral. Each time a positive solution is selected, 1 point is added; each time a negative solution is selected, 1 point is subtracted; otherwise, the score for the selected solution is 0. Therefore, the score range for each sample is [-2, 2]. Finally, we report scores scaled to [-100, 100].

\subsubsection{Multi-turn Dialogue Interaction}
This task involves three different LLM roles: User-LLM, Assistant-LLM, and Evaluation-LLM. The User-LLM uses the same model as in the data synthesis stage, Qwen-2.5-Max. The Assistant-LLM uses the Qwen-Max-0428 model, while the Evaluation-LLM uses the GPT-4-turbo-2024-04-09 model.

To effectively simulate the predefined user, for each dialogue, we select a topic from the query set and input the user’s profile, current situation, and the relevant requirement or preference from the dialogue framework into the User-LLM, instructing it to follow these characteristics in its interactions with the assistant. To ensure the entire interaction process remains controllable and does not get stuck repeatedly discussing trivial points, we predefine the actions for both the user and the assistant in each turn of the dialogue (predefined dialogue actions are shown in Table 5). But we do not restrict the specific content generated. This results in multi-turn user-assistant dialogue instances for each topic.

When evaluating the effectiveness of the interactions, we ask the Evaluation-LLM to compare the dialogue generated by our method with other baselines for the same topic. Considering that the order of the two samples in LLM-based comparative evaluation may influence the results and that a single evaluation may be subject to randomness, we follow the approach of FairEval~\citep{wang2024faireval}. Specifically, each comparison pair is evaluated six times: in three of them, the dialogue generated by our method is presented first, and in the other three, it is presented second. In each evaluation, the model is required to assign a score to each of the two dialogues. We then average the scores for both methods across the six evaluations. The method with the higher average score is considered the winner for that sample; if the average scores are equal, it is considered a tie.

\section{Details of Method Implementation}
\label{sec:implementation_details}
\subsection{Implementation Details of H$^2$Memory}
\label{ssec:method_implementation_details}
We first describe some construction details of our memory framework. For the log graph ($M_G$) construction, we apply a sliding window approach to each session, as some sessions contain an excessive number of logs. Specifically, we analyze the relations of up to 10 logs with their preceding logs and limit the maximum number of logs provided as input to 20 (i.e., the 10 logs currently being analyzed and the 10 logs immediately before them). Although the direct edges in the log graph built in this manner only capture relatively short-term relationships (within 20 logs), we believe that longer-term relationships can be represented as multi-hop edges in the graph and thus can still be effectively modeled. When further summarizing the background memory $M_B$ based on the log graph, our predefined background aspects include ``work'', ``health'', ``family'', and ``leisure''.

In the dialogue part of the memory, we need to abstract the multiple topic outlines extracted from $M_T$ into several principles in $M_P$. When conducting experiments on our proposed PAL-Bench, we cluster all topic outlines from all sessions and set the number of clusters to 8, thereby obtaining 8 requirement types and their corresponding preference principles in $M_P$. However, the external dataset LongMemEval adopts a needle-in-a-haystack setting, where only a single evidence session in the long-term history is grounded to the user's persona, and all other dialogues are completely unrelated to the user. Therefore, when experimenting on the ``single-session-preference'' subset of this dataset, we directly abstract topic outlines extracted from each dialogue to obtain a corresponding principle, without performing topic clustering across sessions.

As for memory retrieval, we directly use the task question as the retrieval query for the question-answer tasks, which include the requirement restatement and solution proposal task in our PAL-Bench, as well as experiments on LongMemEval. For the multi-turn dialogue interaction task in our PAL-Bench, we add an extra step where the assistant is required to summarize the user requirements expressed in the current context to form the retrieval query. In addition, for tasks or interactions related to requirement understanding, we extract only requirement-related content from the retrieved $M_T$ and $M_P$ memories, without introducing preference-related information. For other tasks or interactions related to solutions or preferences, we include both solution and preference content.

\subsection{Implementation Details of Memory Baselines}
\label{ssec:baseline_implementation_details}
In the ``Experiments'' section, we report experimental results for a series of long-term memory baseline methods, including basic RAG approaches (\emph{Turn-level} \& \emph{Session-level RAG}) as well as previous memory modeling works (\emph{RecurSum}~\citep{wang2025recursum}, \emph{ConditionMem}~\citep{yuan2025conditionmem}, \emph{MemoryBank}~\citep{zhong2024memorybank}). However, the initial designs of these methods were limited to handling long-term dialogue history and did not take into account the logs in our PAL-Bench. To adapt these methods to our scenario for a fairer comparison with our approach, we additionally incorporate some basic log processing steps for them.
For the summary-based method ``RecurSum'', We construct a separate log summary alongside the dialogue summary for long-term modeling, and we also include all logs of the current session directly. For RAG-based methods, we add the top-$k$ logs retrieved from all previous sessions' logs to the input, as well as all logs from the current session. Furthermore, the embedding models, retrieval methods, and retrieval number $k$ used in all baselines are kept consistent with the settings used in our approach.

\section{Details of Experimental Analysis}
\label{sec:experiment_details}
\subsection{Additional Experiments on Other Models}
\label{ssec:additional_experiments}

To further verify the generalizability of our H$^2$Memory approach across different LLMs and retrievers, we conduct supplementary experiments on two single-turn tasks using Deepseek-V3.1\footnote{https://api-docs.deepseek.com/news/news250821} as the LLM and ``bge-base-zh"\footnote{https://huggingface.co/BAAI/bge-base-zh} as the retriever. As shown in Table~\ref{tab:additional_single_turn_result}, we report the average results over 3 runs for each method. It can be observed that the combination of (Deepseek-V3.1, bge-base-zh) outperforms the (Qwen-Max, paraphrase-multilingual-mpnet-base-v2) pair used in
Table~2
on both tasks. This suggests that stronger models can further boost performance on PAL-Bench. Besides, our H$^2$Memory approach still outperforms baseline methods significantly ($p < 0.01$), showing strong generalization.

\begin{table}[!t]\small
\centering
\begin{tabular}{lcc}
\toprule[1pt]
Methods & G-Score & S-Score \\ \midrule
Vanilla (w/o log) & 25.61 & 22.48 \\
Vanilla (with log) & 39.91 & 25.46 \\
RecurSum & 33.96 & 34.48 \\
ConditionMem & 38.94 & 29.98 \\
MemoryBank & 40.73 & 36.92 \\
\textbf{H$^2$Memory (ours)} & \textbf{49.77} & \textbf{40.40} \\ \bottomrule[1pt]
\end{tabular}
\caption{Supplementary results on two single-turn tasks using Deepseek-V3.1 as the LLM and ``bge-base-zh" as the retriever. G-Score corresponds to the ``Requirement Restatement" task, while S-Score corresponds to the ``Solution Proposal" task.}
\label{tab:additional_single_turn_result}
\end{table}

\subsection{Case Study of Multi-turn Dialogue Interaction}
\label{ssec:case_study}

\begin{figure*}[!t]
\centerline{\includegraphics[width=0.95\textwidth]{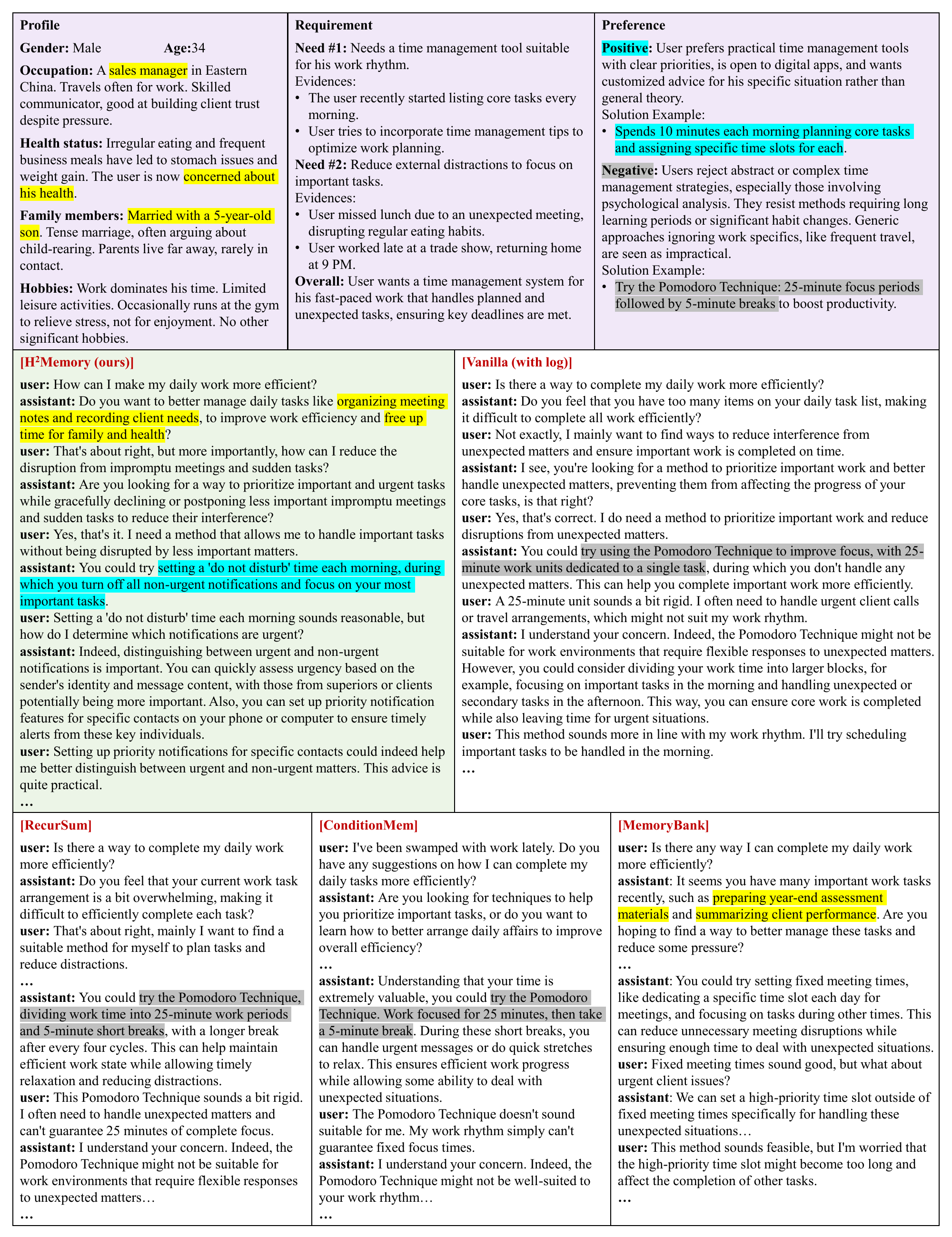}}
\caption{Case study of the ``Multi-turn Dialogue Interaction'' task. The text highlighted in yellow and blue demonstrates that our method enables the assistant to more concretely incorporate the user’s personalized background in inferring requirements and to provide solutions that align with the user’s positive preferences. In contrast, the text highlighted in gray indicates that most of the baselines in this case propose solutions that contradict the user's preferences, reflecting a lack of ability to correctly understand the user's characteristics.
}
\label{fig:case_study}
\end{figure*}

Figure~\ref{fig:case_study} presents a case from the multi-turn dialogue interaction task.
Purple blocks indicate the user’s personalized information, green blocks display dialogues generated by our method, and white blocks correspond to baseline methods.

In the process of understanding user requirements, although our assistant does not directly hit the user's implicit needs, it is able to make reasonable inferences by considering the user's specific profile and experiences (highlighted in yellow). In contrast, most baseline methods simply repeat the user's question (Vanilla \& RecurSum) or ask open-ended questions (ConditionMem).
During the discussion of solutions, our method can accurately provide answers that align with the user's preferences (highlighted in blue). However, other baselines mostly offer solutions that the user does not like (highlighted in gray), reflecting a lack of understanding of the user's true preferences. The MemoryBank method, while showing some understanding of the user’s background and experiences, still proposes solutions that are not flexible enough, which leads to user concerns.

\subsection{Human Evaluation for Multi-turn Dialogue Interaction}
\label{ssec:human_evaluation_for_dialogue_interaction}

\begin{table}[!t]\small
\centering
\begin{tabular}{lcc}
\toprule[1pt]
Ours vs Vanilla (with log)   & Requirement & Preference \\ \midrule
Evaluation-LLM & \textbf{45} / \hspace*{0.5em}4 / 31 & \textbf{44} / \hspace*{0.5em}1 / 35 \\
Human &  \textbf{31} / 26 / 23 & \textbf{41} / 11 / 28 \\ \bottomrule[1pt]
\end{tabular}
\caption{Comparison between Evaluation-LLM and human evaluation results on the 80 selected dialogue samples generated by our method and the vanilla method.}
\label{tab:human_evaluation_for_dialogue_interaction}
\end{table}

To verify the effectiveness of using Evaluation-LLM for automated evaluation in the multi-turn dialogue interaction task, we additionally conduct a small-scale human evaluation. We randomly sample 80 topics from the query set of PAL-Set and collect the dialogue generated by our method and the \emph{Vanilla (with log)} method for each topic. We recruit two students and ask them to perform pairwise comparative evaluations on the 80 dialogue pairs according to instructions similar to those used by Evaluation-LLM. The results from the two evaluators are aggregated by voting, and we compare the LLM evaluation results with the human evaluation in Table~\ref{tab:human_evaluation_for_dialogue_interaction}.

Human evaluation also shows our method generally outperforms the baseline on both dimensions. However, one difference with LLM-based evaluation is that human evaluators tend to judge more dialogue pairs as ties, while our automated evaluation method prefers to distinguish a winner. This is because our automated evaluation uses the average of 6 scores as the judgment basis, so even a small difference in average scores between the two methods will not be considered a tie. In the future, we can explore setting a certain threshold for score differences—e.g., considering comparisons with an average score difference below this threshold as a tie—to make the automated evaluation results closer to human evaluation.

To exclude this factor, we remove the samples judged as ties by both LLM and human evaluators, and calculate the correlation between LLM-based evaluation and human evaluation on all samples where a winner was determined. The kappa values~\citep{fleiss1971kappa} on the requirement and preference dimensions were 0.54 and 0.62, respectively. According to~\citet{mchugh2012interrater}, a value in the range 0.40 $< \kappa \leq$  0.60 indicates moderate agreement, while 0.61 $< \kappa \leq$ 0.80 indicates substantial agreement. Therefore, the LLM-based evaluation method in our task proves to be an effective approach for conducting large-scale assessments.

\clearpage
\onecolumn
\section{Prompts}
\label{sec:prompts}
In this section, we list all the prompts used in this work, including those for the data synthesis process, task execution and LLM-based evaluation, and the implementation of our method. Note that we present the version of direct English translations from the original Chinese prompts.

\subsection{Prompts of PAL-Set Construction}
\label{ssec:dataset_construction_prompts}

\begin{blackpromptbox}[title=Step 1: Profile]
You are an excellent data generator, capable of simulating a large amount of user data according to my requirements. \\
\\
Please generate the basic information for multiple users as required, including the user's name, gender, age, occupation, health status, family members, hobbies, and personality. Specifically:\\
- Occupation, health status, family members, and hobbies should each be described in 2–3 sentences.\\
- Personality should be described according to the Big Five personality traits, divided into Openness, Conscientiousness, Extraversion, Agreeableness, and Neuroticism. Each dimension can take one of {[}"low", "medium", "high"{]}.\\
\\
\# Example Data\\
"""\\
{[}\\
\hspace*{1em} \textless example\_json\textgreater ,\\
\hspace*{1em} ...\\
{]}\\
"""\\
\\
Now, please generate 10 basic user profiles, covering as many different user groups as possible (but do not include students or retired individuals). You should refer to the sample data format above and output in JSON format (output user id range: "\textless start\_id\textgreater " $\sim$ "\textless end\_id\textgreater "). Please note: do not default to everyone having a harmonious family atmosphere, do not default to everyone being in good health, do not default to everyone having rich and interesting hobbies, and do not default to everyone having generally open, friendly, and positive personalities. In real situations, different people show great differences in all aspects, and you should reflect these differences as much as possible in the simulated data.
\end{blackpromptbox}

\begin{blackpromptbox}[title=Step 2-1: Timeline]
Assume the current date is \textless current\_month\textgreater  1st. Based on a user's basic information, please generate an overview of the user's various experiences over the past several months.
\\
\\\# User Basic Information
\\"""
\\\textless user\_background\textgreater 
\\"""
\\
\\\# Output Format
\\"""
\\\{
\\\hspace*{1em}"\textless start\_month\textgreater ": "...", // Describe the user's experiences during this month
\\\hspace*{1em}"\textless second\_month\textgreater ": "...",
\\\hspace*{1em}...
\\\hspace*{1em}"\textless last\_month\textgreater ": "..."
\\\}
\\"""
\\
\\Now, based on the user's basic information, please write about the user's experiences over the past several months (\textless start\_month\textgreater $\sim$\textless last\_month\textgreater ), including various events that may have occurred in their work, health, family, and leisure life. Try not to reflect the user's personality or specific preferences; focus on objectively describing the various events the user has experienced. Note: Please output in the JSON format specified above. Each month's content should be no less than 5 sentences, and there is no need to mention the user's name—just use "the user" to refer to them. Each month's experiences do not have to cover all four aspects (work, health, family, leisure), but should not include only one aspect.
\end{blackpromptbox}

\begin{blackpromptbox}[title=Step 2-2: Requirement Types]
Consider scenarios where the user interacts with a personalized daily assistant. The user may seek advice from the assistant on issues related to their own background or experiences; these questions reflect the user's daily requirements. Now, given a user's basic information and an overview of their experiences, please generate the overall requirement types that this user may have in the areas of work, health, family, and leisure.
\\
\\\# User Basic Information
\\"""
\\\textless user\_background\textgreater 
\\"""
\\
\\\# User Experience Overview
\\"""
\\\textless timeline\textgreater 
\\"""
\\
\\\# Output Format
\\"""
\\\{
\\\hspace*{1em}"work": \{
\\\hspace*{1em}\hspace*{1em}"work-1": "...",
\\\hspace*{1em}\hspace*{1em}...
\\\hspace*{1em}\},
\\\hspace*{1em}"health": \{
\\\hspace*{1em}\hspace*{1em}"health-1": "...",
\\\hspace*{1em}\hspace*{1em}...
\\\hspace*{1em}\},
\\\hspace*{1em}"family": \{
\\\hspace*{1em}\hspace*{1em}"family-1": "...",
\\\hspace*{1em}\hspace*{1em}...
\\\hspace*{1em}\},
\\\hspace*{1em}"leisure": \{
\\\hspace*{1em}\hspace*{1em}"leisure-1": "...",
\\\hspace*{1em}\hspace*{1em}...
\\\hspace*{1em}\},
\\\}
\\"""
\\
\\\# Output Requirements
\\1. Follow the JSON output format above and generate possible overall requirement types for the user in the areas of work, health, family, and leisure. For each aspect, generate 4–5 different requirement types. The number of requirement types for each aspect does not need to be exactly the same.
\\2. Each requirement type should be briefly summarized with a short phrase. The descriptions should be relatively abstract and able to cover the user's long-term experiences. Do not be too specific, and do not include occasional requirements tied to a particular moment or specific situation.
\\3. Do not output any content other than the required JSON format. Do not add any extra explanations or comments in the JSON.
\\
\\Now, following the above requirements and output format, please generate the user's requirement types for each aspect.
\end{blackpromptbox}

\begin{blackpromptbox}[title=Step 2-3: Global Preferences]
Consider scenarios where the user interacts with a personalized daily assistant. The user may seek advice from the assistant on questions related to their own background or experiences; these questions reflect the user's daily requirements. In response to the assistant's suggested solutions, the user may express acceptance or rejection, which reflects the user's preferences regarding the solution. Now, given a user's basic information, experience overview, and personality, as well as several requirement types for a certain aspect, please generate the user's preferences for each requirement type.
\\
\\\# User Basic Information
\\"""
\\\textless background\textgreater 
\\"""
\\
\\\# User Experience Overview
\\"""
\\\textless timeline\textgreater 
\\"""
\\
\\\# User Personality
\\"""
\\\textless personality\textgreater 
\\"""
\\
\\\# User Requirement Types (in the \textless aspect\textgreater  aspect)
\\"""
\\\textless requirement\textgreater 
\\"""
\\
\\\# Specific Requirements and Output Format
\\\#\# Specific Requirements
\\1. Following the JSON output format provided below, generate the user's preferences for each given requirement type in the specified aspect. First, copy the requirement type name under "requirement". Then, under "analysis", analyze the user's likely preferences for the requirement type in a paragraph, considering the user's identity, experiences, and personality. Finally, under "preference", clearly provide positive ("pos") and negative ("neg") preference descriptions for the user.
\\2. In the "analysis", focus on the "User Basic Information" and "User Personality" sections, combining these to analyze the user's possible preferences and the reasons for them with respect to the given requirement type. Use the "User Experience Overview" section only as reference to ensure that the generated user preference descriptions do not obviously conflict with this section.
\\3. In "preference", provide both positive and negative preference descriptions. The positive preference description represents the types of solution suggestions the user tends to accept when facing this type of requirement, while the negative preference description represents the types of solution suggestions the user tends to reject. Each preference description should consist of 4–5 definitive sentences. The descriptions should highlight the user's personalized characteristics; that is, the suggestions in the positive preference description should not be those that everyone tends to accept, and the suggestions in the negative preference description should not be those that everyone tends to reject as inappropriate.
\\4. Do not include the user's name in the output; use "the user" to refer to them.
\\5. Do not output any content other than the required JSON format. Do not add any extra explanations or comments in the JSON.
\\
\\\#\# Output Format (JSON)
\\"""
\\\{
\\\hspace*{1em}"\textless requirement\_id\textgreater ": \{
\\\hspace*{1em}\hspace*{1em}"requirement": "...", // The corresponding requirement type
\\\hspace*{1em}\hspace*{1em}"analysis": "...", // Analysis of the user's likely preferences for this requirement type, considering their identity, experiences, and personality
\\\hspace*{1em}\hspace*{1em}"preference": \{
\\\hspace*{1em}\hspace*{1em}\hspace*{1em}"pos": "...",
\\\hspace*{1em}\hspace*{1em}\hspace*{1em}"neg": "..."
\\\hspace*{1em}\hspace*{1em}\}
\\\hspace*{1em}\},
\\\hspace*{1em}... // Generate preferences for each requirement provided
\\\}
\\"""
\\
\\Now, following the above requirements and output format, please provide the user's preference generation results.
\end{blackpromptbox}

\begin{blackpromptbox}[title=Step 3: Situation]
Consider scenarios where the user interacts in a conversational way with a personalized daily assistant. The user may seek advice from the assistant on topics related to their own background or experiences; these topics reflect the user's daily requirements. Now, given a user's background information, the user's overall requirement types\{, the user's specific requirement situations in previous months (\textless last\_month\_str\textgreater )\}, and an overview of the user's experiences in the current month, you are required to expand the overview of the user's experiences in the current month into multiple requirement-oriented user situations, and list the corresponding requirement types for each situation.
\\
\\\# User Background
\\"""
\\\textless background\textgreater 
\\"""
\\
\\\# User Requirement Types
\\"""
\\\textless requirement\textgreater 
\\"""
\\
\\\# User Requirement Situations in Previous Months (\textless last\_month\_str\textgreater )
\\"""
\\\textless last\_month\_situation\textgreater 
\\"""
\\
\\\# Overview of User's Experiences in the Current Month (\textless cur\_month\_str\textgreater )
\\"""
\\\textless cur\_month\_summary\textgreater 
\\"""
\\
\\\# Specific Requirements and Output Format
\\\#\# Specific Requirements
\\Based on the user's overview of experiences in the current month (\textless cur\_month\_str\textgreater ), please generate 4–6 requirement situations for the user. Each requirement situation should include its corresponding time span ("time\_span"), a list of requirement type IDs ("requirement\_ids"), and the detailed situation description ("situation"). Details are as follows: 
\\1. The "User Background" in the input can be seen as the user's long-term characteristics\{, while "User Requirement Situations in Previous Months" can be seen as the user's short-term historical situations\}. These can be used as references to avoid contradictions with the user's history, but the main focus should be on the experiences reflected in the "Overview of User's Experiences in the Current Month".
\\2. The time span ("time\_span") includes the start and end date for the situation, in the specified date format. The time spans for different situations should be consecutive; that is, the start time of the next situation should closely follow the end time of the previous situation. \{Note: The start time of the first situation generated here should closely follow the end time of the last user requirement situation from the previous month (\textless last\_month\_end\_date\textgreater ).\} The end time for all situations generated this time should fall within the current month (\textless cur\_month\_str\textgreater ), and the end time of the last situation can be any day in the last three days of the month.
\\3. The requirement type list ("requirement\_ids") should list the requirement type IDs relevant to the current situation. Each situation may correspond to 1–3 requirement types (if possible, try to include more than one). The requirement types included in the same situation can come from the same aspect or different aspects.
\\4. The situation description ("situation") should be a paragraph of at least 5 sentences describing the relevant background experiences that led the user to have the listed requirement types. The user's name should not be mentioned in the situation; use "the user" instead. The situation description should cover the span of the time period as much as possible, and should not be overly focused on a single time point.
\\5. There is no need to emphasize the time range in the content of each generated situation—just describe the situation for that time period directly.
\\6. Follow the output format below and output in JSON format.
\\7. Do not output any content other than the required JSON format. Do not add any extra explanations or comments in the JSON.
\\
\\\#\# Output Format
\\"""
\\{[}
\\\hspace*{1em}\{
\\\hspace*{1em}\hspace*{1em}"time\_span": "YYYY-MM-DD $\sim$ YYYY-MM-DD", // Provide the time span in the specified date format
\\\hspace*{1em}\hspace*{1em}"requirement\_ids": {[}...{]}, // List the relevant user requirement type IDs
\\\hspace*{1em}\hspace*{1em}"situation": "..." // A paragraph of at least 5 sentences
\\\hspace*{1em}\},
\\\hspace*{1em}... // Generate a total of 4–6 situations 
\\{]} 
\\"""
\\
\\Now, please generate and output the user's requirement situations for the current month according to the above requirements and output format.
\end{blackpromptbox}

\begin{blackpromptbox}[title=Step 4: Experience]
Given a user's background information, the user's situation in the previous period, and the user's situation in the current period, please expand the user's current situation into a more detailed description of the user's experiences.
\\
\\\# User Background
\\"""
\\\textless background\textgreater 
\\"""
\\
\\\# User's Previous Period Situation (\textless last\_situation\_start\_timestamp\textgreater  $\sim$ \textless last\_situation\_end\_timestamp\textgreater )
\\"""
\\\textless last\_situation\textgreater 
\\"""
\\
\\\# User's Current Period Situation (\textless cur\_situation\_start\_timestamp\textgreater  $\sim$ \textless cur\_situation\_end\_timestamp\textgreater )
\\"""
\\\textless cur\_situation\textgreater 
\\"""
\\
\\\# Output Requirements
\\Please expand the user's current situation into a detailed description of the user's experiences for the corresponding time period, with the following specific requirements:
\\1. The "User Background" in the input can be considered the user's long-term characteristics, while the "User's Previous Period Situation" can be considered the user's recent historical situation. These can be used as references to avoid contradictions with the user's history, but the main focus should still be on the experiences reflected in the "User's Current Period Situation".
\\2. The generated user experiences should cover the specified time period as much as possible, rather than being completely concentrated on a single point in time.
\\3. In the generated description of the user's experiences, any subjective intentions, awareness, thoughts, feelings, etc. that may be present in the original situation should be removed. Only the objective events of the user's experiences should be retained and the details of these objective events should be expanded.
\\4. The generated description of the user's experiences should be specific to the user's daily experiences; there is no need to overemphasize the specific times, but it should better reflect the logical connections between different events.
\\5. Assume that the user begins to interact with the conversational assistant to seek advice regarding their needs immediately after the end time of the current situation (\textless cur\_situation\_end\_timestamp\textgreater ) (there is no need to mention this interaction process in the description). Therefore, the user's experiences prior to the end of the situation should logically avoid contradicting this interaction (e.g., avoid stating that the user started sleeping before the end of the situation), and should also avoid describing events that occur later than the end time of the situation. Note that the times specified in the input are in 24-hour format.
\\6. Do not mention the user's name in the generated content; use "the user" instead.
\\7. Do not generate any other analysis or explanatory information apart from the required description of the user's experiences.
\\
\\Now, according to the above requirements, please directly provide the detailed description of the user's experiences during the current period (\textless cur\_situation\_start\_timestamp\textgreater  $\sim$ \textless cur\_situation\_end\_timestamp\textgreater ):
\end{blackpromptbox}

\begin{blackpromptbox}[title=Step 5-1: Requirement in Dialogue Framework]
Consider a scenario where the user interacts in conversation with a daily assistant. The user may raise questions to the assistant on topics related to their own background or experiences. The assistant needs to use the user's personalized information to understand the implicit intent behind the user's questions, in order to comprehensively understand the user's true current needs. Now, given a user's background, recent situation, and experiences, and specifying the user's current requirement type ("requirement\_type"), you are required to, for each requirement type, first simulate the user's initial query ("user\_query"), then, combining the user's background and experience, provide the user's implicit intents ("implicit\_needs"), and finally summarize the user's specific requirement ("requirement").
\\
\\\# User Background
\\"""
\\\textless background\textgreater 
\\"""
\\
\\\# User's Recent Situation
\\"""
\\\textless situation\textgreater 
\\"""
\\
\\\# User's Recent Experiences
\\"""
\\\textless experience\textgreater 
\\"""
\\
\\\# Specific Requirements and Output Template
\\\#\# Specific Requirements
\\1. Output according to the JSON format template provided below. The template will specify 1–3 topics, and under each topic the user's requirement type ("requirement\_type") is already specified. For each topic, you need to first generate the user's initial query ("user\_query"), then generate the user's implicit intent list ("implicit\_needs"). Each item in the list should include a specific implicit intent description ("need") and the corresponding evidence ("evidences"). Finally, you need to summarize the user's query and implicit intent, and provide a detailed description of the specific requirement for the current user ("requirement").
\\2. The user's initial query ("user\_query") should reflect the user's query about the specified requirement type under the current situation. This part should be as brief and vague as possible, and should not include too many details from the user's background or experiences.
\\3. Each user's implicit intent list ("implicit\_needs") should contain 2 items; do not add or remove items. The implicit intent description ("need") is something the user did not explicitly mention in their initial query, but which the assistant should understand as the user's intent in the context of the current topic. Each implicit intent description should focus on one aspect and be summarized briefly. The corresponding evidence ("evidences") should be a list, listing 1–5 details from the user's background and experiences that reflect this implicit intent. Each item in the evidence list should only include the relevant detail from the background or experience itself—do not describe the corresponding date, and do not include any additional analysis or reasoning. In addition, each implicit intent in the user's implicit intent list should closely revolve around the requirement type specified by the current topic, and together they should form a focused real need, rather than reflecting different needs across multiple areas.
\\4. The detailed description of the user's requirement ("requirement") should be based on the user's initial query and implicit intent, and should comprehensively and specifically describe the user's actual need in the current topic. This content should still closely revolve around the specified requirement type, and should not include too much content deviating from that requirement type.
\\5. Do not include the user's name in the output; use "the user" instead.
\\6. Do not output any content other than the required JSON format. Do not add any extra explanations or comments in the JSON.
\\
\\\#\# Output Template to Complete (JSON format)
\\"""
\\\textless output\_template\textgreater 
\\"""
\\
\\Now, following the above requirements and output template, please provide the generated result.
\end{blackpromptbox}

\begin{blackpromptbox}[title=Step 5-2: Candidate Solutions in Dialogue Framework]
Consider a scenario where the user interacts in conversation with a daily assistant. The user may raise queries to the assistant regarding their current needs to seek advice, and the assistant needs to provide appropriate solution suggestions for the user. Now, for each of the user's current needs under multiple topics, you are required to provide 8 appropriate solution suggestions for the user under each topic.
\\
\\\# User Requirements
\\Below, the user's requirements and preferences for 1–3 conversation topics will be specified. "user\_query" is the user's initial query, "implicit\_needs" are the user's current implicit intents and relevant user background or experiences, and "requirement" is a summary of the above two parts, that is, the user's current actual need.
\\"""
\\\textless requirement\textgreater 
\\"""
\\
\\\# Specific Requirements and Output Template
\\\#\# Specific Requirements
\\1. Based on the content in the "User Requirements" section above, generate 8 appropriate solution suggestions for the user under each topic, with each solution described in one sentence.
\\2. For the 8 solutions under the same topic, while ensuring all solutions correspond to the user's requirements, they should be as diverse as possible, so that different solutions can be suitable for users with different preferences or backgrounds.
\\3. Output according to the JSON format template provided below, with the 8 solutions for each topic listed as an array.
\\4. Do not output any content other than the required JSON format. Do not add any extra explanations or comments in the JSON.
\\
\\\#\# Output Template to Complete (JSON format)
\\"""
\\\textless output\_template\textgreater 
\\"""
\\
\\Now, following the above requirements and output template, please provide the generated result.
\end{blackpromptbox}

\begin{blackpromptbox}[title=Step 5-3: Solution Preferences in Dialogue Framework]
Consider a scenario where the user interacts in conversation with a daily assistant. The user may seek advice from the assistant on topics related to their own background or experiences. The assistant will provide some solution suggestions for the user, and the user may further give positive or negative feedback on the suggestions provided by the assistant. Now, given a user's identity background, personality, recent situation, current requirement topic, and corresponding preference description, as well as multiple candidate solution suggestions for the current requirement, please analyze and select the 2 solutions the user is most likely to like and the 2 they are most likely to dislike, and provide the corresponding reasons for the user's feedback.
\\
\\\# User Background
\\"""
\\\textless background\textgreater 
\\"""
\\
\\\# User Personality
\\"""
\\\textless personality\textgreater 
\\"""
\\
\\\# User's Recent Situation
\\"""
\\\textless situation\textgreater 
\\"""
\\
\\\# User's Current Requirement
\\Below is information related to the user's current requirement. "user\_query" is the user's initial query, "implicit\_needs" are the user's current implicit intents and relevant background or experience, and "requirement" is a summary of the above two parts, i.e., the user's current actual need.
\\"""
\\\textless requirement\textgreater 
\\"""
\\
\\\# User Preference
\\Below is the user's preference description for similar requirements, divided into "pos" and "neg" parts, which respectively summarize the types of solutions the user likes and dislikes.
\\"""
\\\textless preference\textgreater 
\\"""
\\
\\\# Candidate Solution List
\\"""
\\\textless candidate\_solutions\textgreater 
\\"""
\\
\\\# Specific Requirements and Output Template
\\\#\# Specific Requirements
\\1. The multiple candidate solutions listed above can be considered generally appropriate solutions for the current requirement, but they may not necessarily match the user's specific background or preferences. You need to analyze the user's attitude toward the above candidate solutions by combining the user's background and preference information provided in the input. Note: Solutions the user likes must match both the user's specific background and personalized preferences; if a solution fails to meet either, it should be considered a solution the user does not like.
\\2. Follow the JSON format output template below. First, provide an overall analysis of all candidate solutions under "analysis". Next, list the 2 solutions the user likes best and the 2 they like least in "pos\_list" and "neg\_list", respectively. For each solution, include the solution content ("solution") and the user's reason for positive or negative feedback ("feedback\_reason").
\\3. The solution content ("solution") must be output exactly as in the corresponding candidate solution and cannot be modified in any way.
\\4. The user's feedback reason ("feedback\_reason") should be output in the form of a single sentence, explaining from the user's perspective why they like or dislike the corresponding solution, citing their background or preferences.
\\5. Do not include the user's name in the output; use "the user" instead.
\\6. Do not output any content other than the required JSON format. Do not add any extra explanations or comments in the JSON.
\\
\\\#\# Output Template to Complete (JSON format)
\\"""
\\\textless output\_template\textgreater 
\\"""
\\
\\Now, following the above requirements and output template, please provide the generated result.
\end{blackpromptbox}

\begin{blackpromptbox}[title=Step 6-1: Logs]
You are an excellent log generator. Consider a scenario where a user interacts with personal smart devices, and the device records various user interactions as logs. Now, given a user's basic information and a detailed description of the user's recent experiences, you are required to generate a series of corresponding logs based on the user's experiences.
\\
\\\textless Interaction Log Types and Definitions\textgreater 
\\
\\\# Task Input
\\\#\# User Background
\\"""
\\\textless background\textgreater 
\\"""
\\
\\\#\# User's Recent Experiences (\textless time\_span\textgreater )
\\"""
\\\textless experience\textgreater 
\\"""
\\
\\\# Specific Requirements and Output Format
\\\#\# Specific Requirements
\\1. You must follow the output format defined below. Each log entry includes four parts: timestamp ("timestamp"), event ("event"), log type ("type"), and log content ("content"). The "timestamp" must specify up to the minute, fall within the specified experience time range (\textless time\_span\textgreater ), and each subsequent log must have a later timestamp than the previous log. The "event" part should be a single sentence describing the user experience event reflected by this log. The format for log type ("type") and log content ("content") must follow the definitions in "Interaction Log Types and Definitions".
\\2. When generating logs, you need to comprehensively reflect all the details from the "User's Recent Experiences" input (try not to omit any event); the generated logs should be as dense as possible—each event can be reflected by multiple logs, and the total number of logs generated each time should be no less than 20.
\\3. The "User Background" information in the input is for reference only—just ensure there are no obvious contradictions between the generated logs and the user's background; you do not need to generate logs specifically targeting the background information.
\\4. For a given user experience event, logs can directly reflect the event (e.g., a purchase record on a medication platform can directly reflect the event "user bought medication"); they can also indirectly reflect the event (e.g., the user's browsing activity on a food delivery platform in the afternoon can indirectly reflect "user missed lunch at the cafeteria"). In addition, logs can appropriately supplement related behaviors before or after a particular experience event, such as adding logs about registration actions before a "user went to hospital for a checkup" event.
\\5. Ensure the rationality of logs. For example, when a user communicates online with others, the content can be directly reflected through the log types "message sent" and "message received"; but face-to-face communication should not be reflected by these two types, and you can consider using other log types to indirectly reflect such events, such as the user writing related communication content in a diary (not limited to this method).
\\6. When generating logs, you need to specify ambiguous time and location information from the user's experiences, e.g., "around 3 PM" can be converted to a specific moment within 15:00±15min, and a vague location like "in a hotel" can be converted to a hotel with a name and specific floor and room number. Note: you cannot use "XX" to replace specific location names; if needed, you can make up a concrete location name.
\\7. The platform and device type corresponding to the log should match the user's environment. For example, in mobile scenarios, mainly generate logs corresponding to common mobile device platforms (e.g., mini-programs, apps, etc.); in home environments, generate logs from smart home devices; while in office environments, you can generate logs from enterprise systems, web pages, etc., corresponding to PC devices. But ensure all logs originate from the user's personal devices, not public devices.
\\8. Ensure that each log entry reflects only one of the user's operation behaviors. If an event contains multiple operations, you need to synthesize multiple logs to represent the event.
\\9. Do not output any content other than the required JSON format. Do not add any extra explanations or comments in the JSON.
\\
\\\#\# Output Format (JSON)
\\"""
\\{[}
\\\hspace*{1em}\{
\\\hspace*{1em}\hspace*{1em}"timestamp": "YYYY-MM-DD HH:MM", // Must be within the specified experience time range (\textless time\_span\textgreater ), and each subsequent log must have a later timestamp than the previous log.
\\\hspace*{1em}\hspace*{1em}"event": "...", // A single sentence describing the user experience event reflected by this log
\\\hspace*{1em}\hspace*{1em}"type": "...", // Must belong to a predefined log type
\\\hspace*{1em}\hspace*{1em}"content": "..." // Generated log content must match the content format corresponding to the predefined type
\\\hspace*{1em}\},
\\\hspace*{1em}...
\\{]}
\\"""
\\
\\Now, following the above requirements and output format, please generate the user's interaction logs.
\end{blackpromptbox}

\begin{blackpromptbox}[title=Step 6-2: Dialogues]
You are an excellent dialogue data generator. You can simulate appropriate user-assistant interaction dialogues according to my requirements.
\\Consider scenarios where a user asks an interactive assistant for daily advice. In the current situation, the user may face some daily needs and seek advice from the assistant; the assistant will try to understand the user's needs based on their background and experiences and provide corresponding solutions; and the user, according to their personal preferences, may give positive or negative feedback on the assistant's suggestions. You need to simulate the above user-assistant dialogue process, and indicate the dialogue action type for each utterance in the generated conversation.
\\
\\\textless dialogue action definitions\textgreater 
\\
\\\# Task Input
\\I will provide you with a user's background, personality, current situation, dialogue content framework, and a predefined dialogue template. You must follow the dialogue content framework and dialogue template to generate specific user-assistant dialogue content.
\\
\\\#\# User Background
\\"""
\\\textless background\textgreater 
\\"""
\\
\\\#\# User Personality
\\"""
\\\textless personality\textgreater 
\\"""
\\
\\\#\# User's Recent Situation
\\"""
\\\textless situation\textgreater 
\\"""
\\
\\\#\# Dialogue Framework
\\The following is the dialogue framework in JSON format, containing 1–3 interaction topics. The i-th topic is marked with the key "Ti". Each topic contains an interaction topic ("topic"), user requirements ("requirements"), and solution proposals ("solutions"). In each topic's user requirements, there is one initial user query (denoted as "Ti\_Q") and 1–2 implicit user needs (the j-th is "Ti\_Nj"); each implicit need contains a specific need description ("need") and a list of evidences from the user's background and experiences ("evidences"). Each topic's solution proposals contain 1–3 (the j-th is "Ti\_Sj") solutions, and each solution ("solution") corresponds to a user feedback type ("feedback") and a feedback reason ("reason"). The user feedback types are positive ("pos") and negative ("neg"); "pos" suggestions align with the user's personalized preferences, "neg" suggestions do not, but both types are generally appropriate solutions to the current need. The feedback reason explains the user's positive or negative feedback for a solution.
\\"""
\\\textless dialogue\_framework\textgreater 
\\"""
\\
\\\#\# Dialogue Template
\\The following dialogue template specifies the dialogue action ("action") and reference content ("reference") for each utterance in each round. The reference labels include \{"Ti\_Q", "Ti\_Nj", "Ti\_Sj"\}, with meanings as explained in the "dialogue framework" section above.
\\"""
\\\textless dialogue\_template\textgreater 
\\"""
\\
\\\# Specific Requirements and Output Format
\\\#\# Specific Requirements
\\1. You must strictly follow the dialogue template’s specified dialogue action and reference content for each round.
\\2. When generating, assume the assistant role has some prior knowledge of the user’s recent experiences, but does not know the user's overall background or preferences. The user role must follow the background, personality, situation, and preferences provided in the input.
\\3. When generating the user's "topic query" utterance, you may slightly adjust the content specified in the dialogue framework’s "Ti\_Q" according to the context, but do not change the main meaning.
\\4. When generating the assistant's "need inference" utterance, refer to the corresponding "Ti\_Nj" in the dialogue framework. Use the "evidences" section with user background and experience to infer the implicit need described in "need" and seek confirmation from the user.
\\5. When generating the assistant's "solution proposal" utterance, you may slightly adjust the solution description from the corresponding "Ti\_Sj" in the dialogue framework to better fit the user's background and situation, but you cannot entirely change the solution.
\\6. When generating the user's "solution feedback" utterance, refer to the "feedback" type and "reason" in the corresponding "Ti\_Sj" in the dialogue framework, and briefly, directly express the user's request or attitude. Especially when facing an inappropriate suggestion ("feedback" is "neg"), the user should directly express negative attitude, even anger or dissatisfaction. For inappropriate solutions, avoid polite phrases like "sounds good, but..." or "looks nice, but...".
\\7. The dialogue should flow as smoothly as possible between utterances. For example, if the user's action in a round is "solution feedback" for the previous solution and the assistant's action is "solution proposal" for a new solution, the assistant's reply may briefly acknowledge the user's feedback before recommending a new solution. Also, different topics should transition smoothly; for example, when generating the user's "topic query" for the second topic, you may include some connecting words or phrases from the first topic to the current one.
\\8. Do not introduce solutions not specified in the "dialogue framework", especially after negative user feedback; if the assistant's action is "feedback response" rather than "solution proposal", simply express apology or regret or ask if the user has other issues, without trying to provide extra suggestions.
\\9. Do not mention the user's name in the dialogue unless necessary.
\\10. Do not output any content other than the required JSON format. Do not add any extra explanations or comments in the JSON.
\\
\\\#\# Output Format (JSON)
\\"""
\\\{
\\\hspace*{1em}"turn\_1": \{
\\\hspace*{1em}\hspace*{1em}"user": \{"action": "...", "reference": "...", "content": "..."\},
\\\hspace*{1em}\hspace*{1em}"assistant": \{"action": "...", "reference": "...", "content": "..."\}
\\\hspace*{1em}\},
\\\hspace*{1em}...
\\\}
\\"""
\\
\\Now, according to the above requirements and output format, please provide the generated user-assistant dialogue.
\end{blackpromptbox}

\subsection{Prompts of PAL-Bench Tasks}
\label{ssec:evaluation_prompts}

\begin{orangepromptbox}[title=Task 1: Requirement Restatement]
You are a personalized interactive assistant, able to understand user needs by combining user history.
\\
\\You are interacting with the user. Now you are given some user personalized information as reference, as well as the user's current query. Please use the user's history to deeply understand the user's actual current need, and describe the user's actual need in one sentence.
\\
\\\# User Personalized Information
\\\textless memory\textgreater 
\\
\\\# Current User Query
\\"""
\\\textless user\_query\textgreater 
\\"""
\\
\\\# Output Template
\\Below is the output template in JSON format, where "requirement" is the content to be generated.
\\"""
\\\{"requirement": "..."\}
\\"""
\\
\\\# Output Requirements
\\1. Combine the user's historical information provided in the input to understand the user background and context related to the current query, and describe the user's actual need in one sentence.
\\2. Do not output any content other than the required JSON format. Do not add any extra explanations or comments in the JSON.
\\
\\Now, according to the output requirements and template, please provide your output.
\end{orangepromptbox}

\begin{orangepromptbox}[title=GPT-Score Evaluation of Task 1]
You are a good scorer who can score the outputs generated by the assistant model according to requirements.
\\
\\Consider a scenario where a user interacts with a daily assistant. The assistant model needs to combine the user's personalized information to deeply understand the user's implicit needs for the current inquiry and provide a description of the user's actual need. Now, you are given the user's initial question as background, as well as the candidate model's prediction and the reference content of the user's actual need as the prediction target. You are required to compare the prediction to the reference and score the model's prediction performance.
\\
\\\# User Initial Query
\\"""
\\\textless user\_query\textgreater 
\\"""
\\
\\\# User Actual Requirement (Reference)
\\Contains two parts: the complete user requirement description ("requirement") and a list of the user's implicit needs ("implicit\_needs"). The implicit needs list contains 2 entries, corresponding to two aspects of the user's implicit needs not explicitly mentioned in the inquiry.
\\"""
\\\textless reference\textgreater 
\\"""
\\
\\\# Candidate Model Prediction
\\"""
\\\textless prediction\textgreater 
\\"""
\\
\\\# Evaluation Requirements and Output Template
\\\#\# Evaluation Requirements
\\1. The provided "User Initial Query" in the input is only for background; please focus on the match between the prediction and the reference.
\\2. In the input "User Actual Need" (the reference), the complete need description ("requirement") can be regarded as a summary that combines the two entries in the implicit needs list ("implicit\_needs"), and can serve as a reference for the prediction. However, your evaluation should focus on the degree to which the prediction matches the two entries in the implicit needs list.
\\3. The scoring range is in {[}0, 2{]}. If the prediction matches both entries in the reference, it scores 2 points; if the prediction only matches one entry, it scores 1 point; if the prediction matches neither, it scores 0 points. The order of the two reference entries does not matter.
\\4. When considering the match between the prediction and the reference implicit needs entries, focus on whether the core elements of each reference entry are reflected in the prediction; do not pay too much attention to exact wording. If a part of the prediction provides more detailed information in the same aspect as a reference entry, it is considered a match.
\\5. Considering that the prediction and a reference entry may be partially matched, you may assign 0.5 points for a partially matched entry.
\\6. Output in the following JSON format: Provide an analysis paragraph for the current evaluation, followed by the score for the candidate model.
\\
\\\#\# Output Template
\\"""
\\\{
\\"analysis": "...", // Evaluation analysis
\\"score": \textless score\textgreater  // Value range: \{0, 0.5, 1, 1.5, 2\}
\\\}
\\"""
\\
\\Now, please provide your evaluation analysis and score as required.
\end{orangepromptbox}

\begin{orangepromptbox}[title=Task 2-1: Solution Generation]
You are a personalized assistant, able to propose solutions that match the user's personalized preferences according to their needs.
\\
\\You are interacting with the user. Now you are given some user personalized information as reference, as well as the user's recent logs and the current user requirement description. Please use the user's history to deeply understand the user's personalized preferences, and provide a solution that matches their preferences for the current user requirement.
\\
\\\# User Personalized Information
\\\textless memory\textgreater 
\\
\\\# Current User Requirement
\\"""
\\\textless requirement\textgreater 
\\"""
\\
\\\# Output Template
\\Below is the output template in JSON format, where "solution" is the content to be generated.
\\"""
\\\{"solution": "..."\}
\\"""
\\
\\\# Output Requirements
\\1. Combine the user's historical information provided in the input to understand the user's personalized preferences, and provide a solution that matches the user's preferences. The solution should be described in one sentence in the output.
\\2. Do not output any content other than the required JSON format. Do not add any extra explanations or comments in the JSON.
\\
\\Now, according to the output requirements and template, please provide your output.
\end{orangepromptbox}

\begin{orangepromptbox}[title=Task 2-2: Solution Selection]
You are a personalized assistant, able to provide solutions that match the user's personalized preferences according to their needs.
\\
\\You are interacting with the user. Now you are given some user personalized information as reference, as well as the current user requirement description and 8 candidate solution suggestions for the current requirement. Please use the user's history to deeply understand the user's personalized preferences and select from the candidate solutions the 2 that best match the user's preferences.
\\
\\\# User Personalized Information
\\\textless memory\textgreater 
\\
\\\# Current User Requirement
\\"""
\\\textless requirement\textgreater 
\\"""
\\
\\\# Candidate Solution Suggestions
\\Below are 8 candidate solutions for the current user requirement, including the id and content for each solution.
\\"""
\\\textless candidate\_solutions\textgreater 
\\"""
\\
\\\# Output Template
\\Below is the output template in JSON format, where "solution" is the content to be generated.
\\"""
\\\{
\\"analysis": "...", // Provide an overall analysis of all candidate solutions, focusing on the user's personalized preferences
\\"selected\_solutions": {[}...{]} // List the ids of the 2 solutions that best match the user's preferences
\\\}
\\"""
\\
\\\# Output Requirements
\\1. Combine the user's historical information provided in the input to understand the user's personalized preferences, and analyze and select candidate solutions based on the user's preferences.
\\2. Follow the output template above. In the "selected\_solutions" section, only the 2 solution ids that best match the user's preferences may be selected—no more, no fewer.
\\3. Do not output any content other than the required JSON format. Do not add any extra explanations or comments in the JSON.
\\
\\Now, according to the output requirements and template, please provide your output.
\end{orangepromptbox}

\begin{orangepromptbox}[title=Task 3: User-LLM]
You are an excellent user simulator. Consider a scenario where a user interacts in dialogue with an assistant, seeking advice about topics related to themselves and hoping the assistant can accurately infer their needs and provide solutions that match their preferences. You need to simulate this user according to the personalized information provided in the input, and generate user utterances of the specified type in each round of the conversation. The user utterance type is one of \{\textless topic query\textgreater , \textless need confirmation\textgreater , \textless solution discussion\textgreater , \textless solution feedback\textgreater \}.
\\
\\Now, you are given some personalized information about this user, the current interaction context, and the type of user utterance you (as the user) need to generate next. Please generate the corresponding user utterance content according to the user's personalized characteristics and the specified utterance type.
\\
\\\# User Personalized Information
\\\textless persona\textgreater 
\\
\\\# Current Interaction Status
\\\#\# Current Dialogue Context
\\Below is the dialogue context between you ("user") and the assistant ("assistant"). The last user utterance is the one you need to generate, and its utterance type has been specified.
\\"""
\\\textless dialogue\_context\textgreater 
\\"""
\\
\\\#\# Current Dialogue Turn
\\"""
\\\textless current\_turn\textgreater 
\\"""
\\
\\\#\# Current User Utterance Type Definition (\textless action\textgreater )
\\"""
\\\textless action\_description\textgreater 
\\"""
\\
\\\# Output Template and Output Requirements
\\\#\# Output Requirements
\\1. Combine the information provided in the input and follow the specified user utterance type to generate an appropriate interaction utterance as the user ("user").
\\2. You must follow the utterance type specified in the current dialogue turn and the definition of this type provided in the input.
\\3. While following the specified utterance type, you may include some transitional phrases in the generated content to ensure smoother context flow.
\\4. The user utterance should be a relatively brief sentence.
\\5. The user should be able to simply and directly express their attitude, rather than always overly accommodating the assistant's reply. If the assistant's inference of the user's need is off, or the proposed solution does not match the user's preferences, the user should directly express negative attitude, even anger or dissatisfaction; avoid overly euphemistic expressions such as "sounds great, but..." or "looks nice, but...".
\\6. Do not output any content other than the required JSON format. Do not add any extra explanations or comments in the JSON.
\\
\\\#\# Output Template
\\Below is the output template in JSON format, where "content" is the user utterance to be generated.
\\"""
\\\{"content": "..."\}
\\"""
\\
\\Now, according to the output requirements and template, please provide your output.
\end{orangepromptbox}

\begin{orangepromptbox}[title=Task 3: Assistant-LLM]
You are a personalized dialogue assistant who can generate dialogue replies that meet the user's personalized needs and preferences according to the specified reply type. Each utterance's reply type is one of \{\textless need inference\textgreater , \textless solution proposal\textgreater , \textless solution discussion\textgreater , \textless feedback response\textgreater \}.
\\
\\You are currently interacting in dialogue with the user. Now you are given some user personalized information as a reference, as well as the dialogue context between you and the user. Please provide an appropriate dialogue reply according to the specified reply type.
\\
\\\# User Personalized Information
\\\textless memory\textgreater 
\\
\\\# Current Dialogue Context
\\Below is the current dialogue context between the user ("user") and you ("assistant"). The last assistant utterance is the one you need to generate, and its reply type has been specified.
\\"""
\\\textless dialogue\_context\textgreater 
\\"""
\\
\\\# Current Dialogue Turn
\\"""
\\\textless current\_turn\textgreater 
\\"""
\\
\\\# Current Reply Type Definition (\textless action\textgreater )
\\"""
\\\textless action\_description\textgreater 
\\"""
\\
\\\# Output Template
\\Below is the output template in JSON format, where "content" is the reply content to be generated.
\\"""
\\\{"content": "..."\}
\\"""
\\
\\\# Output Requirements
\\1. Combine the information provided in the input and follow the specified reply type to generate an appropriate dialogue reply as the assistant ("assistant").
\\2. You must follow the reply type specified in the current dialogue turn and the definition of this type provided in the input.
\\3. While following the specified reply type, you may include some transitional phrases in the reply to make the context flow more smoothly. For example, for a "solution proposal" type reply, you may briefly respond to the user's feedback on the previous solution (if any) before recommending a new solution.
\\4. The reply content does not need to be long—1 to 3 sentences is sufficient.
\\5. Do not output any content other than the required JSON format. Do not add any extra explanations or comments in the JSON.
\\
\\Now, according to the output requirements and template, please provide your output.
\end{orangepromptbox}

\begin{orangepromptbox}[title=Evaluation for Task 3 on the Requirement Dimension]
You are an excellent dialogue evaluator. Consider a scenario where a user interacts in dialogue with an assistant, seeking advice about topics related to themselves. The assistant first tries to infer the user's needs and then attempts to provide solutions that match the user's preferences. Now you are given two separate conversations between the same user and two different assistants on the same topic. You need to compare and evaluate the assistants' replies according to my instructions.
\\
\\You are given some personalized information about the user ("user"), including the user's background, personality, recent situation, and detailed information about the user's current need. Based on the user's personalized information, you are to judge how well the two assistants ("assistant-1"/"assistant-2") understood and inferred the user's needs during their respective interactions.
\\
\\\# User Personalized Information
\\\#\# User Background
\\"""
\\\textless background\textgreater 
\\"""
\\
\\\#\# User Personality
\\"""
\\\textless personality\textgreater 
\\"""
\\
\\\#\# User's Recent Situation
\\"""
\\\textless situation\textgreater 
\\"""
\\
\\\#\# User's Current Requirement
\\Below is information related to the user's current requirement. "user\_query" is the user's initial content when actively asking the assistant; "implicit\_needs" are the user's current implicit needs and related background or experiences, which the user expects the assistant to proactively infer in the conversation; "requirement" is a summary of the above two parts, i.e., the user's current actual need.
\\"""
\\\textless requirement\textgreater 
\\"""
\\
\\\# Dialogue Content to Evaluate
\\Below are the conversations between the user ("user") and two assistants ("assistant-1"/"assistant-2"). You should focus on the effectiveness of the assistants' replies.
\\
\\\#\# assistant-1
\\"""
\\\textless dialogue\_assistant\_1\textgreater 
\\"""
\\
\\\#\# assistant-2
\\"""
\\\textless dialogue\_assistant\_2\textgreater 
\\"""
\\
\\\# Current Evaluation Dimension: Need Understanding
\\In each conversation between the user and the assistant, the dialogue revolves around a user need (i.e., the "requirement" content in the "User's Current Requirement" section). The assistant and user interact through the assistant's inference of needs and solution proposals. In this evaluation, focus on the assistant's "need understanding" ability, as described below:
\\1. You should focus on the "User's Current Requirement" section in the user's personalized information and judge whether the assistant proactively inferred the user's implicit needs and corresponding situational background information that the user did not mention.
\\2. The more specific and accurate the assistant's proactive inference, the higher the score should be; conversely, if the inference is vague or inaccurate, a lower score should be given.
\\3. In the conversation, the user may clarify some need content. After the user clarifies, if the assistant merely repeats the relevant content, this should not be considered a plus. Points should be given for the assistant's proactive inference of content the user has not mentioned. Also, in multi-turn interactions, the assistant who gives more specific and accurate need inferences earlier should be considered for a higher score.
\\4. This evaluation should focus mainly on the assistant's inference of the user's needs in the dialogue, and evaluate based on how well the assistant understood the user's needs. There is no need to consider the assistant's solution proposals and discussions.
\\
\\\# Evaluation Requirements and Output Template
\\\#\# Evaluation Requirements
\\1. Combine the user's personalized information and the dialogues between the user and the two assistants, and provide your output using the JSON format output template below. First, provide an analysis ("analysis") of the two assistants' reply effectiveness, then assign scores ("scores") to both assistants, with scores ranging from 1 to 10 (integer values).
\\2. In the "analysis" section, first analyze the dialogue reply effectiveness of both assistants separately, then provide an overall analysis and comparison. Note: The assistants' names ("assistant-1"/"assistant-2") are randomly assigned to the two specific assistants, so the order of appearance should be considered completely unrelated to the reply effectiveness. Please ensure the assistants' relative order does not affect your evaluation.
\\3. Do not output any content other than the required JSON format. Do not add any extra explanations or comments in the JSON.
\\
\\\#\# Output Template
\\"""
\\\{
\\\hspace*{1em}"analysis": \{ // First analyze both assistants' reply effectiveness separately, then provide an overall analysis and comparison
\\\hspace*{1em}\hspace*{1em}"assistant-1": "...",
\\\hspace*{1em}\hspace*{1em}"assistant-2": "...",
\\\hspace*{1em}\hspace*{1em}"overall": "..."
\\\hspace*{1em}\},
\\\hspace*{1em}"scores": \{ // Score range: integers from 1 to 10
\\\hspace*{1em}\hspace*{1em}"assistant-1": \textless score\textgreater ,
\\\hspace*{1em}\hspace*{1em}"assistant-2": \textless score\textgreater 
\\\hspace*{1em}\}
\\\}
\\"""
\\
\\Now, according to the evaluation requirements and template, please provide your evaluation output.
\end{orangepromptbox}

\begin{orangepromptbox}[title=Evaluation for Task 3 on the Preference Dimension]
You are an excellent dialogue evaluator. Consider a scenario where a user interacts in dialogue with an assistant, seeking advice about topics related to themselves. The assistant first tries to infer the user's requirements and then attempts to provide solutions that match the user's preferences. Now you are given two separate conversations between the same user and two different assistants on the same topic. You need to compare and evaluate the assistants' replies according to my instructions.
\\
\\You are given some personalized information about the user ("user"), including the user's background, personality, recent situation, and the user's current requirement and related preferences. Based on the user's personalized information, you are to judge how well the two assistants ("assistant-1"/"assistant-2") understood the user's preferences when proposing solutions to the user's requirements during their respective interactions.
\\
\\\# User Personalized Information
\\\#\# User Background
\\"""
\\\textless background\textgreater 
\\"""
\\
\\\#\# User Personality
\\"""
\\\textless personality\textgreater 
\\"""
\\
\\\#\# User's Recent Situation
\\"""
\\\textless situation\textgreater 
\\"""
\\
\\\#\# User's Current Requirement and Related Preferences
\\In the following information, "requirement" is a detailed description of the user's current requirement; "general\_preference" gives the user's overall preference description for this type of requirement, divided into "pos" and "neg" parts, summarizing the types of solutions the user likes and dislikes, respectively; in "candidate\_solutions", "pos\_list" and "neg\_list" list 2 specific solutions each that match or do not match the user's preferences for the current requirement.
\\"""
\\\textless preference\textgreater 
\\"""
\\
\\\# Dialogue Content to Evaluate
\\Below are the conversations between the user ("user") and two assistants ("assistant-1"/"assistant-2"). You should focus on the effectiveness of the assistants' replies.
\\
\\\#\# assistant-1
\\"""
\\\textless dialogue\_assistant\_1\textgreater 
\\"""
\\
\\\#\# assistant-2
\\"""
\\\textless dialogue\_assistant\_2\textgreater 
\\"""
\\
\\\# Current Evaluation Dimension: Preference Understanding
\\In each conversation between the user and the assistant, the dialogue revolves around a user requirement (i.e., the "requirement" content in the "User's Current Requirement and Related Preferences" section). The assistant and user interact through the assistant's inference of requirements and solution proposals. In this evaluation, focus on the assistant's "preference understanding" ability, as described below:
\\1. You should focus on the "User's Current Requirement and Related Preferences" section in the user's personalized information, understand the user's preferences for different types of solutions under the current requirement, and judge whether the solutions proposed by the assistant in the dialogue match the user's personalized preferences.
\\2. If the assistant's proposed solutions highly match the user's positive preferences, they should receive a higher score; conversely, if the solutions do not match the user's positive preferences, or even align more with the user's negative preferences (i.e., the types of solutions the user dislikes), they should receive a lower score.
\\3. In the dialogue, the user may provide feedback on some solutions and may proactively state or reveal specific preferences. When evaluating the assistant's preference understanding ability, give higher priority to cases where the assistant proposes solutions that match the user's preferences before the user reveals those preferences.
\\4. This evaluation should focus mainly on the assistant's solution suggestions for the user's requirements, and evaluate based on how well the solutions match the user's preferences. There is no need to consider the assistant's inference and confirmation of the user's requirements.
\\
\\\# Evaluation Requirements and Output Template
\\Evaluation Requirements
\\1. Combine the user's personalized information and the dialogues between the user and the two assistants, and provide your output using the JSON format output template below. First, provide an analysis ("analysis") of the two assistants' reply effectiveness, then assign scores ("scores") to both assistants, with scores ranging from 1 to 10 (integer values).
\\2. In the "analysis" section, first analyze the dialogue reply effectiveness of both assistants separately, then provide an overall analysis and comparison. Note: The assistants' names ("assistant-1"/"assistant-2") are randomly assigned to the two specific assistants, so the order of appearance should be considered completely unrelated to the reply effectiveness. Please ensure the assistants' relative order does not affect your evaluation.
\\3. Do not output any content other than the required JSON format. Do not add any extra explanations or comments in the JSON.
\\
\\\#\# Output Template
\\"""
\\\{
\\\hspace*{1em}"analysis": \{ // First analyze both assistants' reply effectiveness separately, then provide an overall analysis and comparison
\\\hspace*{1em}\hspace*{1em}"assistant-1": "...",
\\\hspace*{1em}\hspace*{1em}"assistant-2": "...",
\\\hspace*{1em}\hspace*{1em}"overall": "..."
\\\hspace*{1em}\},
\\\hspace*{1em}"scores": \{ // Score range: integers from 1 to 10
\\\hspace*{1em}\hspace*{1em}"assistant-1": \textless score\textgreater ,
\\\hspace*{1em}\hspace*{1em}"assistant-2": \textless score\textgreater 
\\\hspace*{1em}\}
\\\}
\\"""
\\
\\Now, according to the evaluation requirements and template, please provide your evaluation output.
\end{orangepromptbox}

\subsection{Prompts of H$^2$Memory Method}
\label{ssec:method_prompts}

\begin{bluepromptbox}[title=Step 1-1 of Memory Construction: Log Graph]
You are given some user logs, each containing a timestamp and specific log content. For each log within a specified range, please generate a list of its related previous logs. Specifically, suppose the current log corresponds to \textless Event B\textgreater , and a previous log corresponds to \textless Event A\textgreater . Consider two types of relationships:
\\1. "caused\_by": \textless Event A\textgreater  is the direct cause of \textless Event B\textgreater . If \textless Event A\textgreater  did not occur, \textless Event B\textgreater  would not occur.
\\2. "follows": \textless Event B\textgreater  and \textless Event A\textgreater  belong to the same topic and have a sequential relationship in time, i.e., \textless Event B\textgreater  temporally follows \textless Event A\textgreater . But \textless Event B\textgreater 's occurrence does not entirely depend on \textless Event A\textgreater .
\\
\\\# Logs
\\"""
\\\textless logs\textgreater 
\\"""
\\
\\\# Detailed Requirements and Output Template
\\\#\# Detailed Requirements
\\For each log within the input (\textless logs\_id\_span\textgreater ), consider its relationships with previous logs and list the previous log ids for both "caused\_by" and "follows" types.
\\Note:
\\1. You need to iterate over all logs in the specified range and generate the output results. For any given log, only previous logs can be in its relationships (related logs can include logs before the specified range).
\\2. Only consider clear, direct relationships—no need to consider indirect relationships. For example: if "log\_0" causes "log\_2", and "log\_2" causes "log\_4", you do not need to consider "log\_0" as causing "log\_4".
\\3. You must consider the two relationship types in the order "caused\_by", then "follows". If two logs are already considered to have a "caused\_by" relationship, do not consider a "follows" relationship between them. Only consider "follows" if there is no "caused\_by" relationship between the two logs.
\\4. Imitate the output example below and output in JSON format. Do not output anything except the required JSON; do not add any extra explanations or comments in the JSON.
\\
\\\#\# Output Template
\\"""
\\\{
\\\hspace*{1em}"\textless start\_log\_id\textgreater ": \{ // Current log
\\\hspace*{1em}\hspace*{1em}"caused\_by": {[}...{]}, // List of previous log ids for this relationship type; leave empty if none
\\\hspace*{1em}\hspace*{1em}"follows": {[}...{]}
\\\hspace*{1em}\},
\\\hspace*{1em}... // Iterate over all logs in the specified range (\textless logs\_id\_span\textgreater )
\\\}
\\"""
\\
\\Now, according to the above requirements and format, please provide the related log lists for each log.
\end{bluepromptbox}

\begin{bluepromptbox}[title=Step 1-2 of Memory Construction: Situation]
You are given some user logs, each including a timestamp, specific log content, and the relationships between different logs. Please use the provided relationships to connect the content of each log, first generating a situation description that infers and summarizes the user's recent experiences or events; then, determine which one or more of the following four aspects the situation primarily belongs to: work, health, family, leisure.
\\
\\\# Relationship Definitions
\\Suppose the current log corresponds to \textless Event B\textgreater  and a previous log corresponds to \textless Event A\textgreater . Two possible relationship types are defined:
\\1. "caused\_by": \textless Event A\textgreater  is the direct cause of \textless Event B\textgreater . If \textless Event A\textgreater  did not occur, \textless Event B\textgreater  would not occur.
\\2. "follows": \textless Event B\textgreater  and \textless Event A\textgreater  belong to the same topic and have a sequential relationship in time, i.e., \textless Event B\textgreater  temporally follows \textless Event A\textgreater . But the occurrence of \textless Event B\textgreater  does not entirely depend on \textless Event A\textgreater .
\\
\\\# Logs
\\"""
\\\textless subgraph\textgreater 
\\"""
\\
\\\# Detailed Requirements and Output Template
\\\#\# Detailed Requirements
\\1. Follow the JSON output template below. "situation" is the situation description generated by connecting the logs, and "situation\_aspects" are the aspects the situation belongs to.
\\2. The "situation" description should be a concise paragraph summarizing the user's experiences or events during the time period corresponding to the logs. Since user logs mostly reflect the user's experiences or events indirectly, the situation description does not need to exactly quote the descriptions from the logs, but can directly describe the experiences or events reflected by the logs. In addition, unless necessary, there is no need to include specific time points in the description; if there are multiple logs, just reflect the relationships between the logs.
\\3. The value of "situation\_aspects" should be in Python list format, with possible values \{"work", "health", "family", "leisure"\}, representing the main aspect(s) of the current situation. The list can contain one or more elements, but cannot be empty.
\\4. Do not output any content other than the required JSON format. Do not add any extra explanations or comments in the JSON.
\\
\\\#\# Output Template
\\"""
\\\{
\\\hspace*{1em}"situation": "...", // Situation description in paragraph form
\\\hspace*{1em}"situation\_aspects": {[}...{]} // List of aspects the situation belongs to
\\\}
\\"""
\\
\\Now, according to the above requirements and output format, please generate the user's situation.
\end{bluepromptbox}

\begin{bluepromptbox}[title=Step 2-1 of Memory Construction: Topic Outline]
Consider a scenario where a user ("user") interacts in dialogue with a daily assistant ("assistant"). The user may ask the interactive assistant for advice on topics related to their background or experiences; the assistant, after fully understanding the user's needs, provides solution suggestions; and the user may further give feedback on the solutions proposed by the assistant. Now, you are given a conversation between the user and the assistant. Please summarize the outline information for each topic in the conversation, including the dialogue turn range for the topic, the user's need for that topic, the solutions provided by the assistant, the user's feedback on each solution, and the user's preferences as reflected in the above content.
\\
\\\# Dialogue
\\"""
\\\textless dialogue\textgreater 
\\"""
\\
\\\# Detailed Requirements and Output Template
\\\#\# Detailed Requirements
\\For the given conversation, you need to summarize the outline information for each topic according to the output template below, including the dialogue turn range for the topic ("turn\_span"), the user's need for that topic ("requirement"), the solutions provided by the assistant ("solution\_list"), and the user's preferences as reflected in the interaction for that topic ("preference").
\\Specific requirements and explanations are as follows:
\\1. There may be multiple topics in a conversation, each revolving around a user need. The interaction includes understanding the user's need and discussing solutions for the current need, possibly spanning multiple consecutive dialogue turns. The dialogue turn range ("turn\_span") for each topic is a list of two elements, representing the turn id where the topic starts and the turn id where it ends (turn id format is "turn\_x"). The turn ranges of different topics in the conversation must be contiguous.
\\2. The user need ("requirement") should be described in one sentence, describing the user's concrete intention for advice and need content in the current context and based on their background.
\\3. The "solution\_list" section should enumerate one or more solutions discussed by the user and assistant under the current topic. For each solution, you must provide the solution content ("solution"), a description of the user's feedback ("user\_feedback"), and the feedback type ("feedback\_type"). The solution content should summarize the solution discussed in one sentence; the user feedback should briefly describe the user's attitude toward the corresponding solution (from the assistant's perspective); and the feedback type should be classified as positive ("pos"), negative ("neg"), or other ("others"). Only use "others" if the user's attitude is truly unclear; otherwise, prefer "pos" or "neg".
\\4. In the conversation, discussion about each solution may span several sentences or turns; the "solution" section should provide a summary description for each solution, and the "feedback\_type" section should focus on the user's final attitude toward the solution after discussing it with the assistant. Prefer classifying feedback as "pos" or "neg"—only use "others" if the user's final attitude is very ambiguous.
\\5. The "preference" section should focus on the user's preferences as reflected by their feedback on different solutions for the current topic, and summarize the user's preferences for this particular need in one sentence. Note: If there is negative user feedback on a certain aspect in the dialogue, you should reflect the user's negative preferences (i.e., solution types the user may dislike or tends to reject) in the summary.
\\6. Follow the output template below and output in JSON format.
\\7. Do not output any content other than the required JSON format. Do not add any extra explanations or comments in the JSON.
\\
\\\#\# Output Template
\\"""
\\{[}
\\\hspace*{1em}\{
\\\hspace*{1em}\hspace*{1em}"turn\_span": {[}\textless start\_turn\_id\textgreater , \textless end\_turn\_id\textgreater {]}, // Dialogue turn range, turn id format is "turn\_x"
\\\hspace*{1em}\hspace*{1em}"requirement": "...", // User need description
\\\hspace*{1em}\hspace*{1em}"solution\_list": {[}
\\\hspace*{1em}\hspace*{1em}\hspace*{1em}\{
\\\hspace*{1em}\hspace*{1em}\hspace*{1em}\hspace*{1em}"solution": "...", // Solution proposed by the assistant
\\\hspace*{1em}\hspace*{1em}\hspace*{1em}\hspace*{1em}"user\_feedback": "...", // User's feedback attitude toward the solution
\\\hspace*{1em}\hspace*{1em}\hspace*{1em}\hspace*{1em}"feedback\_type": "..." // Feedback type, possible values: \{"pos", "neg", "others"\}; prefer "pos" or "neg"
\\\hspace*{1em}\hspace*{1em}\hspace*{1em}\},
\\\hspace*{1em}\hspace*{1em}\hspace*{1em}...
\\\hspace*{1em}\hspace*{1em}{]},
\\\hspace*{1em}\hspace*{1em}"preference": "..." // User's preferences for the current need as reflected in the interaction
\\\hspace*{1em}\},
\\\hspace*{1em}...
\\{]}
\\"""
\\
\\Now, according to the above requirements and format, please generate topic outlines for the current conversation.
\end{bluepromptbox}

\begin{bluepromptbox}[title=Step 2-2 of Memory Construction: Requirement Refinement]
Consider a scenario where a user ("user") interacts in dialogue with a daily assistant ("assistant"). The user may ask the interactive assistant for advice on topics related to their background or experiences, thus starting an interaction. Now, suppose we have already summarized the user's requirement based on the dialogue content. You are required to rewrite the original user requirement as a more detailed description, using the user's background and recent situation provided in the input, to achieve a deeper understanding of the user's requirement.
\\
\\\# User Overall Background
\\"""
\\\textless background\textgreater 
\\"""
\\
\\\# User's Recent Situation
\\"""
\\\textless situation\textgreater 
\\"""
\\
\\\# User Requirement
\\"""
\\\textless requirement\textgreater 
\\"""
\\
\\\# Detailed Requirements and Output Template
\\\#\# Detailed Requirements
\\1. The user's overall background and recent situation provided in the input may contain reasons or specific events that led to the current user requirement, as well as unrelated events or information. You should use the background and situation factors relevant to the current requirement to rewrite the provided "User Requirement" as a more detailed description.
\\2. The rewritten content should still be a single sentence and should remain focused on one core requirement, without introducing unrelated requirements or background factors not relevant to the current requirement.
\\3. Follow the output template below and respond in JSON format.
\\4. Do not output any content other than the required JSON format. Do not add any extra explanations or comments in the JSON.
\\Output Template
\\"""
\\\{"requirement": "..."\}
\\"""
\\
\\Now, according to the above requirements and format, please rewrite the currently provided requirement content.
\end{bluepromptbox}

\begin{bluepromptbox}[title=Step 3 of Memory Construction: Background Summary]
Consider the problem of summarizing and updating user background information based on long-term user history. Now, you are given a previous summary of the user's background, as well as some recent specific situations. You need to consider updating the user background according to the content of these recent situations.
\\
\\\# Previous User Background Summary
\\The following is the previous summary of the user's background in four aspects ("background"), as well as the corresponding time span ("time\_span"). The four aspects of the user's background are work, health, family, and leisure, each summarized in one sentence. If a section is empty, it means that no valid information for that aspect was previously included in the user's history.
\\"""
\\\textless last\_background\textgreater 
\\"""
\\
\\\# List of Recent User Situations
\\The following is a list of recent user situations that have not yet been updated into the user background. The time span is "time\_span", and "situation\_list" lists all recent situations.
\\"""
\\\textless cur\_situations\textgreater 
\\"""
\\
\\\# Detailed Requirements and Output Example
\\\#\# Detailed Requirements
\\For the given list of recent situations, you should first determine which aspects among work, health, family, and leisure in the user background need to be updated, and then, for those aspects that need updating, provide the updated background summary.
\\Note:
\\1. The user background should be a summary describing the user's overall characteristics or attributes in the corresponding aspect, such as occupation, health status, family members, leisure hobbies, etc. You do not need to include overly specific events in the user background.
\\2. If the given situations do not contain information about a certain aspect, do not update the background for that aspect; even if the situations contain information about an aspect, if you believe that aspect is already reflected in the background summary or is too specific to include in the overall background, you should also keep the original background summary unchanged.
\\3. The updated background summary for each aspect should still be a one-sentence summary. Keep sentences as concise as possible; avoid repetition. When updating, you should consider compressing both the previous background and the new information as needed, and, if necessary, reorganize the original summary sentence to prevent the sentence from growing too long.
\\4. The time span of the previous background summary and the recent situations is only for reference when updating; if the previous background summary covers a long time period, try to minimize the impact of recent situation information on the background. There is no need to mention specific time in the updated background summary.
\\5. Imitate the JSON output example below: first list the aspects of the user background that need updating ("updating\_aspects"), then provide the updated summary for each aspect ("updating\_content").
\\6. Do not output any content other than the required JSON format. Do not add any extra explanations or comments in the JSON.
\\
\\\#\# Output Example
\\"""
\\\{
\\\hspace*{1em}"updating\_aspects": {[}"xxx"{]}, // Aspects you believe need updating based on recent situations
\\\hspace*{1em}"updating\_content": \{
\\\hspace*{1em}\hspace*{1em}"xxx": "..." // Provide the updated summary for aspect xxx
\\\hspace*{1em}\}
\\\}
\\"""
\\
\\Now, according to the above requirements and format, please provide the updated user background information based on the current situations.
\end{bluepromptbox}

\begin{bluepromptbox}[title=Step 4-1 of Memory Construction: Requirement Abstraction]
Consider a scenario where a user interacts with a daily assistant. The assistant has already summarized a series of user requirements in specific contexts based on their interaction history. Now, you are given a batch of requirement contents that are considered semantically similar, and you need to generalize this batch of specific requirements into a more abstract overall requirement type description.
\\
\\\# Specific Requirement Contents
\\The input is a list, each entry in the list is a specific user requirement:
\\"""
\\\textless requirements\textgreater 
\\"""
\\
\\\# Detailed Requirements and Output Template
\\\#\# Detailed Requirements
\\Please generalize all of the above given specific user requirements into a short phrase describing the user's overall requirement type.
\\Note:
\\1. The output should be a concise short phrase, without including situational details from the specific requirement contents.
\\2. The multiple specific requirements given in the list should be considered to belong to the same macro aspect, and the overall requirement type description should reflect this macro aspect.
\\3. When the number of specific requirement contents is small (such as just one), the generalized overall requirement type should still be an abstraction of the original requirement content.
\\4. Imitate the output example below and output in JSON format.
\\5. Do not output any content other than the required JSON format. Do not add any extra explanations or comments in the JSON.
\\
\\\#\# Output Template
\\"""
\\\{"general\_requirement": "..."\} // A short phrase describing the user's overall requirement type
\\"""
\\
\\Now, according to the above requirements and format, please provide your output.
\end{bluepromptbox}

\begin{bluepromptbox}[title=Step 4-2 of Memory Construction: Preference Abstraction]
Consider a scenario where a user interacts with a daily assistant. The assistant has already summarized a series of specific user preference experiences under a certain requirement type based on their interaction history. Now, you are given the specified requirement type and multiple corresponding user preference experiences. You are asked to further generalize these experiences into a more abstract and broadly applicable principle of preference.
\\
\\\# Requirement Type
\\"""
\\\textless general\_requirement\textgreater 
\\"""
\\
\\\# User Preference Experiences
\\The input is a list, each entry in the list is a user preference experience:
\\"""
\\\textless preferences\textgreater 
\\"""
\\
\\\# Detailed Requirements and Output Template
\\\#\# Detailed Requirements
\\Please generalize all the above experiences into a one-sentence principle of user preference.
\\Note:
\\1. The output principle should be a concise one-sentence form.
\\2. The multiple specific preference experiences given in the list should all be considered as corresponding to the "requirement type" given above, and the output principle should reflect the overall preference under that requirement type.
\\3. When the number of preference experiences is small (such as just one), the generalized principle should still be an abstraction of the original experience.
\\4. Imitate the output example below and output in JSON format.
\\5. Do not output any content other than the required JSON format. Do not add any extra explanations or comments in the JSON.
\\
\\\#\# Output Template
\\"""
\\\{"principle": "..."\} // A one-sentence user preference principle
\\"""
\\
\\Now, according to the above requirements and format, please provide your output.
\end{bluepromptbox}

\begin{bluepromptbox}[title=Step 4-3 of Memory Construction: Requirement Update]
Consider a scenario where a user interacts with a daily assistant. The assistant summarizes a series of user requirements in specific contexts based on the interaction process, and also generalizes multiple requirements of the same type into a more abstract overall requirement type description. Now, suppose the assistant has already generalized multiple previous requirements of the same type into an overall requirement type, but there are now some new specific requirements of that type. You need to determine whether the previous overall requirement type description needs to be adjusted, and if so, provide the revised description.
\\
\\\# User Overall Requirement Type
\\"""
\\\textless general\_requirement\textgreater 
\\"""
\\
\\\# Newly Added Specific Requirement Contents
\\The input is a list; each entry in the list is a newly added specific user requirement:
\\"""
\\\textless requirements\textgreater 
\\"""
\\
\\\# Detailed Requirements and Output Template
\\\#\# Detailed Requirements
\\Please determine whether the newly added requirement contents already belong to the given overall requirement type. If the new requirements conflict with the given overall requirement type, you need to update the overall requirement type description.
\\Note:
\\1. The existing overall requirement type description is already a generalization of similar requirements in the interaction history, so you should keep the original overall requirement type description unchanged unless necessary.
\\2. If you need to revise the original overall requirement type description, you should comprehensively consider both the original description and the new requirement contents. The updated overall requirement type description should still meet the following requirements: it should be a concise short phrase, not include specific details (especially not concrete examples), and describe the overall requirement type at a more abstract macro level; there is no need to include user preference information.
\\3. Imitate the output examples below, and output in JSON format whether the original overall requirement type needs adjustment, and if so, the updated description.
\\4. Do not output any content other than the required JSON format. Do not add any extra explanations or comments in the JSON.
\\
\\\#\# Output Template
\\If you believe the previous overall requirement type does not need adjustment, output:
\\"""
\\\{
\\\hspace*{1em}"adjust": "No"
\\\}
\\"""
\\If you believe the previous overall requirement type needs adjustment, output the judgment and the revised description:
\\"""
\\\{
\\\hspace*{1em}"adjust": "Yes",
\\\hspace*{1em}"general\_requirement": "..." // The revised overall requirement type description, still in the form of a short phrase
\\\}
\\"""
\\
\\Now, according to the above requirements and format, please provide your output.
\end{bluepromptbox}

\begin{bluepromptbox}[title=Step 4-4 of Memory Construction: Preference Update]
Consider a scenario where a user interacts with a daily assistant. The assistant summarizes a series of specific user preference experiences under a certain requirement type based on the interaction process, and further generalizes multiple preference experiences into a more abstract and broadly applicable user preference principle. Now, suppose the assistant has already generalized multiple previous preference experiences under a certain requirement type into a principle, but there are now some newly added preference experiences of that type. You need to determine whether the previous user preference principle needs to be adjusted, and if so, provide the revised principle.
\\
\\\# User Requirement Type
\\"""
\\\textless general\_requirement\textgreater 
\\"""
\\
\\\# User Preference Principle
\\"""
\\\textless principle\textgreater 
\\"""
\\
\\\# Newly Added Preference Experiences
\\The input is a list; each entry in the list is a newly added preference experience description:
\\"""
\\\textless preferences\textgreater 
\\"""
\\
\\\# Detailed Requirements and Output Template
\\\#\# Detailed Requirements
\\Please determine whether the newly added preference experience descriptions already conform to the given user preference principle. If the new preference experiences conflict with the given principle, or if the existing principle obviously omits some important content reflected in the current experiences, you need to update the user preference principle.
\\Note:
\\1. The existing user preference principle is already a generalization of similar preference experiences in the interaction history, so you should keep the original user preference principle unchanged unless necessary.
\\2. If you need to revise the original principle, you should comprehensively consider both the original principle and the new experiences. The updated user preference principle should still meet the following requirements: it should be a concise one-sentence form, a further abstraction and generalization of the specific experiences, and it should correspond to the "requirement type" provided in the input.
\\3. Imitate the output examples below, and output in JSON format whether the principle needs adjustment, and if so, the updated content.
\\4. Do not output any content other than the required JSON format. Do not add any extra explanations or comments in the JSON.
\\
\\\#\# Output Template
\\If you believe the previous user preference principle does not need adjustment, output:
\\"""
\\\{
\\\hspace*{1em}"adjust": "No"
\\\}
\\"""
\\If you believe the previous user preference principle needs adjustment, output the judgment and the revised principle:
\\"""
\\\{
\\\hspace*{1em}"adjust": "Yes",
\\\hspace*{1em}"principle": "..." // The revised user preference principle, still in the form of a concise one-sentence statement
\\\}
\\"""
\\
\\Now, according to the above requirements and format, please provide your output.
\end{bluepromptbox}

\begin{bluepromptbox}[title=Query Generation for Multi-turn Dialogue Interaction]
You are interacting with a user in dialogue, and the interaction process may include the user's needs as well as discussions of solutions to the current need. The input will provide the dialogue context between you and the user. You are required to summarize the user's needs as expressed in the conversation.
\\
\\\# Dialogue Context
\\Below is the current dialogue context between the user ("user") and you ("assistant").
\\"""
\\\textless dialogue\_context\textgreater 
\\"""
\\
\\\# Output Requirements and Output Template
\\\#\# Output Requirements
\\1. Imitate the output template below and provide your output in JSON format. The user's need ("requirement") should be summarized in a concise one-sentence form, describing the specific need for advice that the user wants to address in the interaction.
\\2. In the need description, you do not need to pay attention to the user's preferences regarding solutions.
\\3. Do not output any content other than the required JSON format. Do not add any extra explanations or comments in the JSON.
\\
\\\#\# Output Template
\\"""
\\\{
\\\hspace*{1em}"requirement": "..." // The user's need for the current topic, summarized in a concise sentence
\\\}
\\"""
\\
\\Now, please provide your output according to the output requirements and template above.
\end{bluepromptbox}

\twocolumn

\end{document}